\documentclass[10pt,twocolumn,letterpaper]{article}

\usepackage[pagenumbers]{cvpr} 

\usepackage{times}
\usepackage{epsfig}
\usepackage{graphicx}
\usepackage{amsmath}
\usepackage{amssymb}

\usepackage{bm}
\usepackage{enumerate}
\usepackage{arydshln}
\usepackage{booktabs}
\usepackage{multirow}
\usepackage{tabularx}
\usepackage{algorithmic}
\usepackage{algorithm}
\usepackage{amsthm}
\usepackage{color}
\usepackage{xfrac}
\usepackage{threeparttable}
\usepackage{siunitx}
\usepackage{caption}
\usepackage{hhline}
\usepackage[table]{xcolor}
\usepackage{array}
\usepackage{threeparttable}
\usepackage[accsupp]{axessibility}  

\definecolor{myGreen}{RGB}{46,204,113}
\definecolor{myRed}{RGB}{255,0,0}
\definecolor{myYellow}{RGB}{255,255,0}
\definecolor{myBlue}{RGB}{52,152,219}

\definecolor{second}{RGB}{255, 255, 200}
\definecolor{best}{RGB}{255, 220, 200}

\newcolumntype{C}{>{\centering\arraybackslash}X}

\usepackage[pagebackref,breaklinks,colorlinks]{hyperref}
\usepackage[symbol]{footmisc}

\usepackage[capitalize]{cleveref}
\crefname{section}{Sec.}{Secs.}
\Crefname{section}{Section}{Sections}
\Crefname{table}{Table}{Tables}
\crefname{table}{Tab.}{Tabs.}

\def\cvprPaperID{5174} 
\def\confName{CVPR}
\def\confYear{2022}

\begin{document}

\title{HDR-NeRF: High Dynamic Range Neural Radiance Fields}

\author{Xin Huang$^{1}$\footnote{}, Qi Zhang$^{2}$, Ying Feng$^{2}$, Hongdong Li$^{3}$, Xuan Wang$^2$, Qing Wang$^1$ \vspace{6pt}\\
$^{1}$ School of Computer Science, Northwestern Polytechnical University, Xi'an 710072, China \\
$^{2}$ Tencent AI Lab \qquad $^{3}$ Australian National University\\
{\tt\small xinhuang@mail.nwpu.edu.cn} \qquad 
{\tt\small \{nwpuqzhang, yfeng.von, xwang.cv\}@gmail.com}\\
{\tt\small HONGDONG.LI@anu.edu.au} \qquad 
{\tt\small qwang@nwpu.edu.cn}
}



\twocolumn[{%
\renewcommand\twocolumn[1][]{#1}%
\maketitle

\begin{center}
    \centering
    \captionsetup{type=figure}
    \subfloat[Input views]{\includegraphics[width=0.3\linewidth]{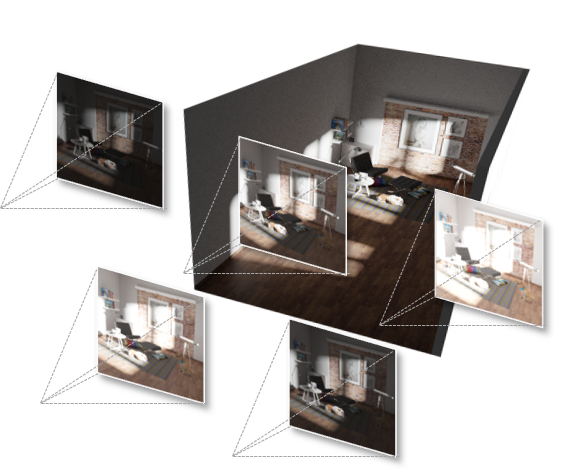}%
    \label{fig:overview_a}}
    \hfil 
    \subfloat[Novel LDR views]{\includegraphics[width=0.345\linewidth]{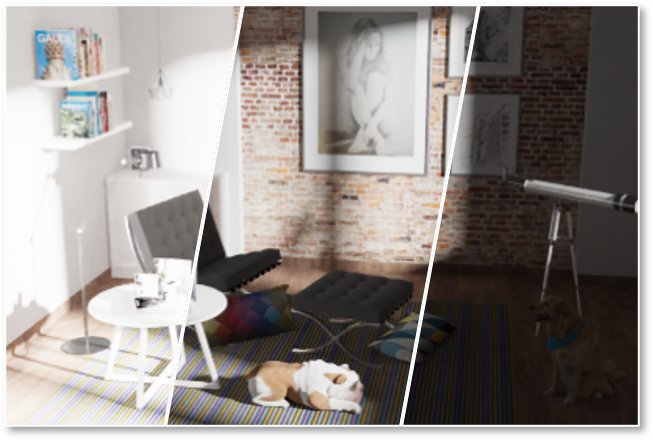}%
    \label{fig:overview_b}}
    \hfil
    \subfloat[One of the novel HDR views]{\includegraphics[width=0.34\linewidth]{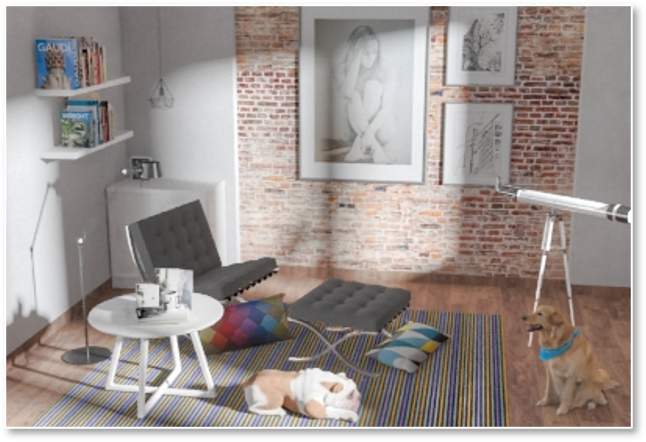}
    \label{fig:overview_c}}
    \caption{We recover a high dynamic range neural radiance field from (a) multiple LDR views with different exposures. Our system is able to render (b) novel LDR views with arbitrary exposures and (c) novel HDR views.}
    \label{fig:overview}
\end{center}%
}]

\footnotetext[1]{Work done during an internship at Tencent AI Lab.}

\begin{abstract}
We present High Dynamic Range Neural Radiance Fields (HDR-NeRF) to recover an HDR radiance field from a set of low dynamic range (LDR) views with different exposures. Using the HDR-NeRF, we are able to generate both novel HDR views and novel LDR views under different exposures. The key to our method is to model the simplified physical imaging process, which dictates that the radiance of a scene point transforms to a pixel value in the LDR image with two implicit functions: a radiance field and a tone mapper. The radiance field encodes the scene radiance (values vary from $0$ to $+\infty$), which outputs the density and radiance of a ray by giving corresponding ray origin and ray direction. The tone mapper models the mapping process that a ray hitting on the camera sensor becomes a pixel value. The color of the ray is predicted by feeding the radiance and the corresponding exposure time into the tone mapper. We use the classic volume rendering technique to project the output radiance, colors and densities into HDR and LDR images, while only the input LDR images are used as the supervision. We collect a new forward-facing HDR dataset to evaluate the proposed method. Experimental results on synthetic and real-world scenes validate that our method can not only accurately control the exposures of synthesized views but also render views with a high dynamic range. 
\end{abstract}

\section{Introduction}
Novel view synthesis is one of the most pursued topics in computer graphics and computer vision. Limited by the dynamic range of camera sensors and input views, rendered novel views are often with a low dynamic range, while human eyes are able to perceive a much higher dynamic range than what is possible by a regular camera. It is therefore highly desirable to render novel HDR views to improve the overall visual experience.


Recently, a series of works have been focused on recovering the radiance field of a scene to render photorealistic novel views using deep neural networks \cite{mildenhall2020nerf,yu2021pixelnerf,martin2021nerf,barron2021mip}. They implicitly encode volumetric densities and colors using a multi-layer perceptron (MLP), which is termed {\em neural radiance field} (NeRF). These methods produce high-quality novel views, yet the dynamic range of the obtained radiance in NeRF is limited to a low dynamic range (between $0$ and $255$), while the radiance in the physical world scene often covers a much broader (higher) dynamics range (\eg from $0$ to $+\infty$). We also notice that `NeRF in the Dark' tries to recover radiance field from raw images with noise \cite{mildenhall2021nerf}, while it's different from our method.

High Dynamic Range (HDR) imaging is the set of techniques that recover HDR images from multiple LDR images with different exposures \cite{szeliski2010computer}. The most common way to reconstruct HDR images is to take a series of LDR images with different exposures at a fixed camera pose and then merge those LDR images into an HDR image \cite{debevec1997recovering, mertens2007exposure,reinhard2010high}. These methods produce compelling results for tripod-mounted cameras but may lead to ghost artifacts when the camera is hand-held. To overcome the limitations of conventional multi-exposure stack-based HDR synthesis, some deep learning methods have been proposed to solve this problem via a two-stage approach \cite{kalantari2017deep,yan2019attention}: 1) aligning the input LDR images using optical flow or removing plausible motion regions, 2) merging the processed images into an HDR image. However, in cases with large motion, their approach typically introduces artifacts in the final results. Most critically, these HDR imaging methods are unable to render novel views and the learning-based methods require HDR images as training supervision. To render novel views, some methods try to merge image-based rendering and HDR imaging techniques. \cite{ruckert2021adop, lechlek2019interactive, sharma2012parameterized, lu2010multi}. However, the image-based methods struggle from preserving view consistency.

\begin{figure}[tb]
    \centering
    \includegraphics[width=\hsize]{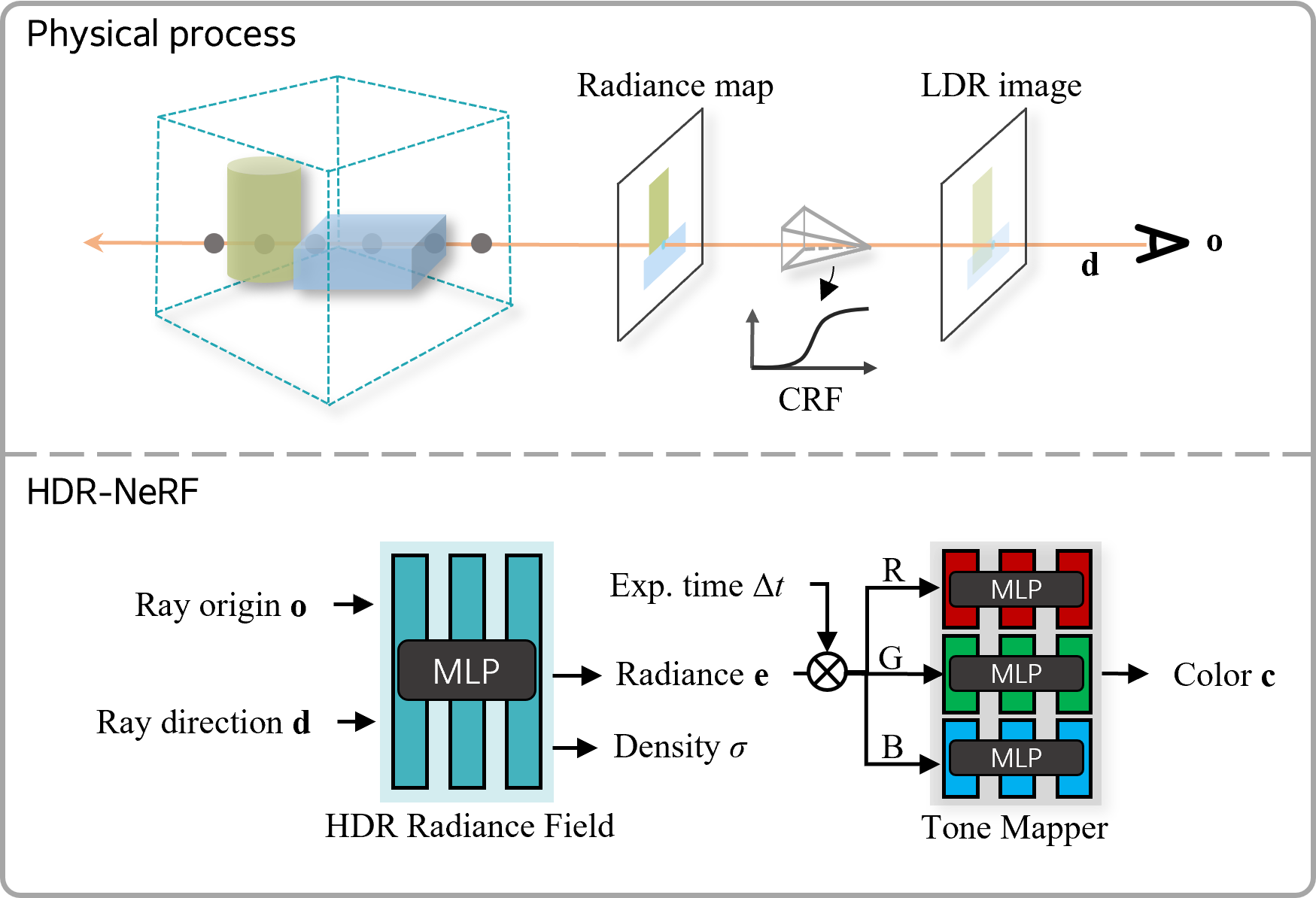}
    \caption{The pipeline of HDR-NeRF modeling the simplified physical process. Our method is consisted of two modules: an HDR radiance field models the scene for radiance and densities and a tone mapper models the CRF for colors.}
    \label{fig:pipeline}
\end{figure}

In this paper, we propose a method HDR-NeRF to recover the high dynamic range neural radiance field from a set of LDR images (\cref{fig:overview_a}) with various \textit{exposures} (the exposure is defined as the product of exposure time and radiance). To the best of our knowledge, this is the first end-to-end neural rendering system that can render novel HDR views (\cref{fig:overview_c}) and control the exposure of novel LDR views (\cref{fig:overview_b}). Building upon NeRF, we introduce a differentiable tone mapper to model the process that radiance in the scene becomes pixel values in the image. 
We use an MLP to model the tone-mapping operation. 
Overall, HDR-NeRF can be represented by two continuous implicit neural functions: a \textit{radiance field} for density and scene radiance and a \textit{tone mapper} for color, as shown in \cref{fig:pipeline}. 
Our pipeline enables joint learning of the two implicit functions, which is critical to recovering the HDR radiance field from such sparse sampled LDR images. We use the classical volume rendering technique \cite{kajiya1984ray} to accumulate radiance, colors, and densities into HDR and LDR images, but we only use LDR ground truth as supervision.

To evaluate our method, we collect a new HDR dataset that contains synthetic scenes and real-world scenes. We compare our method with original NeRF \cite{mildenhall2020nerf}, NeRF-W (NeRF in the wild) \cite{martin2021nerf}, as well as NeRF-GT (a version of NeRF that is trained from LDR images with consistent exposures or HDR images). We provide quantitative and qualitative results and ablation studies to justify our main technical contributions. Our method achieves similar scores across all major metrics on this dataset compared with NeRF-GT. Besides, compared to the recent state-of-the-art NeRF and NeRF-W, our method can render LDR novel views with arbitrary exposures and spectacular novel HDR views.
The main contributions of this paper can be summarized as follows:

\begin{enumerate}
    \item An end-to-end method HDR-NeRF is proposed to recover the high dynamic range neural radiance field from multiple LDR views with different amounts of exposure. 
    \item The camera response function is modeled, both HDR views and LDR views with varying exposures are rendered from the radiance field. 
    \item A new HDR dataset including synthetic and real-world scenes is collected. Compared with SOTAs, our method achieves the best performance on this dataset. The dataset and code will be released for further research purposes in this community. 
\end{enumerate}





\section{Related Work}
\noindent\textbf{Novel View Synthesis.}  Novel view synthesis aims to generate novel images from a new viewpoint using a set of input views.  It is a typical application of image-based rendering technique ~\cite{shum2000review}, such as rendering novel views using depth \cite{zitnick2004high, chaurasia2013depth, zhou2013plane, cayon2015bayesian, overbeck2018system} or explicit geometry information \cite{debevec1996modeling, yu20143d, hedman2017casual, hedman2018instant}. Many classic IBR methods estimate radiance of input images using HDR imaging methods to render novel HDR views \cite{ruckert2021adop, lechlek2019interactive, sharma2012parameterized, lu2010multi}. The estimated radiance using HDR imaging methods is always image-wise. It may be hard to preserve the view consistency in challenging scenes.  On the other hand, light field rendering methods interpolate views based on implicit soft geometry estimates derived from densely sampled images \cite{mcmillan1995plenoptic, levoy1996light, gortler1996lumigraph, buehler2001unstructured, davis2012unstructured}.

In recent years, deep learning techniques have been applied to novel view synthesis to get high-quality photorealistic views. These learning-based approaches can be classified into three categories according to scene representation models.  The first category aims to combine the convolutional neural network (CNN) with traditional voxel grid representation \cite{lombardi2019neural,sitzmann2019deepvoxels,chen2020neural}, such that Sitzmann \textit{et al}. \cite{sitzmann2019deepvoxels} use a CNN to compensate the discretization artifacts from low resolution voxel grids. Lombardi \textit{et al}. \cite{lombardi2019neural} control the predicted voxel grids based on the input time of dynamic scene. Inspired by the layered depth images \cite{shade1998layered}, other learning-based methods focus on training a CNN to predict a multi-plane images representation from a set of input images and render novel views using alpha-compositing \cite{zhou2018stereo, choi2019extreme, mildenhall2019local,flynn2019deepview}. These methods predict multi-planes images to synthesize views for specific applications, such as light-field rendering \cite{mildenhall2019local} and baseline magnification\cite{zhou2018stereo}. The third category is the NeRF family which represents a scene with a neural radiance field \cite{mildenhall2020nerf,yu2021pixelnerf,martin2021nerf,barron2021mip,boss2021nerd,li2021neural,ma2021deblur,chen2021hallucinated}. Although these recent methods achieved high-quality of rendered novel views, none of them has tackled the task of synthesizing a novel view with {\em high dynamic range}.


\noindent\textbf{Neural Implicit Representation.} Recently, there has been a surge in representing 3D scenes in implicit functions via a neural network. Compared to traditional explicit representations, such as point cloud \cite{pumarola2020c}, voxels \cite{girdhar2016learning} and octrees \cite{wang2017cnn}, neural implicit representations have shown high-quality view synthesis results such as continuous and high-fidelity. We focus on the neural radiance fields representation that implicitly models the volume densities and colors of the scenes with MLPs \cite{mildenhall2020nerf}. NeRF approximates a continuous 3D function by mapping from an input 5D location to scene properties. Recently, NeRF has been explored for novel view relighting \cite{boss2021nerd,srinivasan2021nerv}, view synthesis for dynamic scenes \cite{li2021neural,xian2021space,pumarola2021d,du2021neural,park2021nerfies,li2021neural3d}, scene editing \cite{guo2020object,zhang2021editable,Yang_2021_ICCV,martin2021nerf}. Particularly, Martin-Brualla \textit{et al}. \cite{martin2021nerf} propose NeRF-W to build NeRF from internet photo collections with different photometric variations and occlusions. They learn a per-image latent embedding to capture photometric appearance variations in training images, which enable them to modify the lighting and appearance of a rendering. Although various extensions have been explored to NeRF, which enables them to effectively represent the scene radiance captured by cameras. However, all the NeRF based methods ignore the physics process from radiance to pixel values, which hinders them from representing the radiance in the real world.

\noindent\textbf{High Dynamic Range Imaging.} Traditional multiple exposures-based HDR imaging methods reconstruct HDR images by calibrating the CRF from an exposure stack that a series of LDR images under different exposures with a same pose \cite{debevec1997recovering} or directly merge the LDR images into an HDR image \cite{mertens2007exposure}. To overcome the limitations of traditional methods, such as ghosting in the HDR results when LDR images are captured by a hand-held camera or on a dynamic scene, some methods are proposed to detect the motion regions in the LDR images and then remove these regions in the fusion \cite{grosch2006fast,jacobs2008automatic}. In contrast, alignment-based methods align the input multiple LDR images by estimating optical flow then merge the aligned images \cite{tursun2015state,kalantari2017deep,yan2019robust}. Depending on the great potential of deep learning, some methods try to reconstruct an HDR image from a single LDR image \cite{eilertsen2017hdr,khan2019fhdr,kim2021end}. However, most HDR imaging methods require the given LDR images with a fixed or quasi-fixed camera pose. Besides, these methods can only synthesize HDR images with original poses and require ground truth HDR images to supervise.





\section{Background}\label{sec:background}
\subsection{Neural Radiance Fields}

NeRF \cite{mildenhall2020nerf} represents a scene using an implicit neural function, which maps a ray origin $\mathbf{o} = (x, y, z)$ and ray direction $\mathbf{d}=(\theta,\phi)$ into a color $\mathbf{c}=(r,g,b)$ and density $\sigma$, that is $(\mathbf{o},\mathbf{d})\rightarrow(\mathbf{c},\sigma)$.
Specifically, suppose a camera ray $\mathbf{r}$ is emitted from camera center $\mathbf{o}$ with direction $\mathbf{d}$, \ie $\mathbf{r}(s)=\mathbf{o} + s\mathbf{d}$ where $s$ denotes a position along the ray. The expected color $\widehat{C}(\mathbf{r})$ of $\mathbf{r}(s)$ is defined as:

\begin{equation}\small
	\widehat{C}(\mathbf{r})=\int_{s_n}^{s_f}T(s)\sigma(\mathbf{r}(s))\mathbf{c}(\mathbf{r}(s), \mathbf{d}) \ ds,
	\label{eq:nerf}
\end{equation}

\begin{equation}\small
	T(s)=\mathrm{exp}\left( - \int_{s_n}^{s}\sigma(\mathbf{r}(p)) \ dp \right),
	\label{eq:occlusion}
\end{equation}
where $s_n$ and $s_f$ denote the near and far boundary of the ray respectively, and $T(s)$ denotes an accumulated transmittance. The predicted pixel value is then compared to the ground truth $C(\mathbf{r})$ for optimization. For all the camera rays of the target view with a pose $\mathbf{P}$, the color reconstruction loss is thus defined by
\begin{equation}\small
	\mathcal{L} = \sum_{\mathbf{r}\in{\mathcal{R}(\mathbf{P})}} \| \widehat{C}(\mathbf{r}) - C(\mathbf{r}) \|^2,
	\label{eq:nerf_loss}
\end{equation}
where $\mathcal{R}(\mathbf{P})$ is a set of camera rays at target position $\mathbf{P}$.

In practice, naively feeding 5D coordinates into the MLP results in renderings that struggle from representing high-frequency variation in color and geometry. To tackle this problem, a positional encoding strategy is adopted in NeRF. Besides, NeRF simultaneously optimizes two models, where the densities predicted by the coarse model are used to bias the sample of a ray in the fine model.

\subsection{Camera Response Functions}

In most imaging devices, the incoming irradiance is mapped into pixel values and stored in images by a series of linear and nonlinear image processing (\eg white balance). In general, all the image processing can be combined in a single function $f$ called \textit{camera response function} (CRF) \cite{dufaux2016high}. It's hard to know the CRFs of cameras beforehand, because they are intentionally designed by the camera manufacturers. Taking ISO gain and aperture as implicit factors, without loss of generality, the nonlinear mapping can be modeled as \cite{szeliski2010computer}: 
\begin{equation}\small
	Z = f(H\Delta t),
	\label{eq:CRF}
\end{equation}
where $H$ is irradiance, the total amount of light incident on a camera sensor, $Z$ denotes the pixel value, and $\Delta t$ denotes exposure time which is decided by the shutter speed. Note that, in the neural radiance field, the integration of scene radiance over the lens aperture is ignored and the irradiance is considered as radiance \cite{dufaux2016high}. 

\section{HDR Neural Radiance Fields}\label{sec:method}
In this section, we introduce our method HDR-NeRF for recovering high dynamic range neural radiance fields. As shown in \cref{fig:pipeline}, our method consists of two main modules to be described in this section. Our goal is to recover the real radiance field in which the radiance is between $0$ and $+\infty$ by using the LDR images with different exposures as supervision. The main challenge is how to efficiently aggregate information in the LDR images to get an HDR radiance field.

\subsection{Scene Representation}
To render novel HDR views, we represent the scene as an HDR radiance field within a bounded 3D volume. An MLP $F$ called \textit{radiance field} is used to model the HDR scene radiance, which is similar to NeRF. For a given ray origin $\mathbf{o}$ and ray direction $\mathbf{d}$, the \textit{radiance field} $F$ outputs the radiance $\mathbf{e}$ and density $\sigma$ of the ray $\mathbf{r}(s) = \mathbf{o}+s\mathbf{d}$, which is formulated as:

\begin{equation}\small
    \left( \mathbf{e}(\mathbf{r}), \sigma(\mathbf{r}) \right) = F(\mathbf{r}).
    \label{eq:radiance_field}
\end{equation}
Note that, the outputs of implicit function in NeRF are colors and densities, while our outputs are radiance and densities.

\subsection{Learned Tone-mapping}
Representing a scene with an HDR radiance field, the key is how to ensure \textit{radiance field} outputs the radiance of ray without the HDR ground truth as supervision. Inspired by the CRF calibration that the process of determining the mapping between the digital value of a pixel and the corresponding irradiance (up to a scale factor), a \textit{tone mapper} is introduced to model the nonlinear mapping of HDR rays to LDR rays. Specifically, we use an MLP $f$ to estimate the CRF of a camera and map our predicted radiance into colors. According to \cref{eq:CRF}, our predicted radiance $ \mathbf{e}$ 
by \cref{eq:radiance_field} is then tone-mapped into color $ \mathbf{c}$. We formulate the differentiable tone-mapping operation as: 

\begin{equation}\small
    \textbf{c}(\mathbf{r}, \Delta t) = f(\mathbf{e}(\mathbf{r})\Delta t(\mathbf{r})), 
    \label{eq:tone_mapper}
\end{equation}
where $\Delta t(\mathbf{r})$ denotes the exposure time of a camera for capturing the ray $\mathbf{r}$. We can easily read exposure time from the EXIF files that contain metadata about photos, such as exposure time, focal length, f-number, \etc. 
In practice, the RGB channels of images are tone-mapped with different CRFs, hence three MLPs are used in our method to process each channel independently.


Following the classical nonparametric CRF calibration method by Debevec and Malik \cite{debevec1997recovering}, we transform all the images into a logarithm radiance domain to optimize the network. Specifically, we assume the \textit{tone mapper} $f$ is monotonic and invertible, so we can rewrite \cref{eq:tone_mapper} as:
\begin{equation}\small
    \ln f^{-1}\left(\textbf{c}(\mathbf{r} ,\Delta t)\right) = \ln \mathbf{e}(\mathbf{r}) + \ln \Delta t(\mathbf{r}).
	\label{eq:radiance}
\end{equation}
We then present the inverse function of $\ln f^{-1}$ as $g$, thus: 
\begin{equation}\small
    \textbf{c}(\mathbf{r}, \Delta t) = g \left( \ln \mathbf{e}(\mathbf{r}) + \ln \Delta t(\mathbf{r}) \right),
	\label{eq:inverse}
\end{equation}
where $g= (\ln{{f^{-1}}})^{-1} $. As a result, our \textit{tone mapper} function is transformed to function $g$ with a logarithm radiance domain.

\subsection{Neural Rendering}
We use the conventional volume rendering technique \cite{kajiya1984ray} to render the color of each ray passing through the scene. Combining the \textit{radiance field} module and \textit{tone mapper} module, we substitute \cref{eq:inverse} into \cref{eq:nerf}. The expected color $\widehat{C}(\mathbf{r}, \Delta t)$ of ray $\mathbf{r}(s)$ with near and far bounds $s_n$ and $s_f$ is given by:

\begin{equation}\small
    \widehat{C}(\mathbf{r}, \Delta t) = \int_{s_n}^{s_f}T(s)\sigma(\mathbf{r}(s))g \left( \ln \mathbf{e}(\mathbf{r}(s)) + \ln \Delta t(\mathbf{r}) \right) \ ds,
	\label{eq:render_ldr}
\end{equation}
where $T(s)$ is defined in \cref{eq:occlusion}. To render HDR views, the tone-mapping operation is removed. Similarly, an HDR pixel value is approximated as:

\begin{equation}\small
    \widehat{E}(\mathbf{r}) = \int_{s_n}^{s_f}T(s)\sigma(\mathbf{r}(s))\mathbf{e}(\mathbf{r}(s)) \ ds.  
	\label{eq:render_hdr}
\end{equation}





\subsection{Optimization}

\noindent\textbf{Color reconstruction loss.} To optimize the two implicit functions $F$ and $g$ from input LDR images, we minimize the mean squared error (MSE) between the LDR views rendered by HDR-NeRF and the ground truth LDR views. Similar to NeRF, we simultaneously optimize a coarse model and a fine model. The color reconstruction loss is formulated as:

\begin{equation}\small
	\mathcal{L}_{c} \!=\! \sum_{\mathbf{r}\in{\mathcal{R}(\mathbf{P})}} \| \widehat{C}_c(\mathbf{r}, \Delta t) - C(\mathbf{r}, \Delta t) \|^2_2 + \| \widehat{C}_f(\mathbf{r}, \Delta t) - C(\mathbf{r}, \Delta t) \|^2_2,
	\label{eq:loss_color}
\end{equation}
where $C$ is the ground-truth color of each pixel, and $\widehat{C}_c$ and $\widehat{C}_f$ are the color predicted by the coarse model and fine model respectively.

\begin{table*}[th]
    \small
    \centering
    \caption{Quantitative comparisons with baseline methods on synthetic and real scenes. Metrics are averaged over the scenes from our dataset (per-scene metrics are shown in supplementary material). LDR-OE denotes the LDR results with exposure $t_1$, $t_3$, and $t_5$. LDR-NE denotes the LDR results with exposure $t_2$, and $t_4$. HDR denotes the HDR results. We color code each column as \colorbox{best}{best} and \colorbox{second}{second best}.}
    
    \begin{threeparttable}
    \begin{tabular}{@{}p{60pt}|p{30pt}|p{30pt}<{\centering} p{30pt}<{\centering} p{30pt}<{\centering} p{30pt}<{\centering} p{30pt}<{\centering} p{30pt}<{\centering} p{30pt}<{\centering} p{30pt}<{\centering} p{30pt}<{\centering}}
        \hline 
        & & \multicolumn{3}{c}{LDR-OE ($t_1,t_3,t_5$)} 
        & \multicolumn{3}{c}{LDR-NE ($t_2,t_4$)}
        & \multicolumn{3}{c}{HDR} \\
        \cline{3-11} 
    &   & PSNR$\uparrow$ & SSIM$\uparrow$ & LPIPS$\downarrow$ & PSNR$\uparrow$ & SSIM$\uparrow$ & LPIPS$\downarrow$ & PSNR$\uparrow$ & SSIM$\uparrow$ & LPIPS$\downarrow$ \\
        \hline


\multirow{2}*{NeRF\cite{mildenhall2020nerf}} 
& Syn.  & 13.97 	& 0.555  &0.376 & ---  & ---  & ---  & --- & --- & --- \\ 
& Real  & 14.95   & 0.661   & 0.308  & ---     & ---     &  ---      &  ---      & ---   &  ---    \\

        \hline
\multirow{2}*{NeRF-W\tnote{1} \cite{martin2021nerf}} 
& Syn.  & 29.83 	& 0.936  & 0.047  & 29.22 	& 0.927 	& 0.050 
& ---  & ---   & ---   \\
& Real  & 28.55   & 0.927   & 0.094  & 28.64   & 0.923   & 0.089  & --- & --- & ---  \\
        \hline
\multirow{2}*{NeRF-GT\tnote{2} \cite{mildenhall2020nerf}} 
& Syn.  &\cellcolor{second}37.66   &\cellcolor{second}0.965   &\cellcolor{second}0.028    &\cellcolor{second}35.87  &\cellcolor{second}0.955   &\cellcolor{second}0.032  &\cellcolor{best}37.80 &\cellcolor{best}0.964  &\cellcolor{best}0.029 \\
& Real  &\cellcolor{best}34.55   &\cellcolor{best}0.958   &\cellcolor{best}0.057   &\cellcolor{best}34.59   &\cellcolor{best}0.956   &\cellcolor{best}0.051  & --- & --- & ---   \\
\hline
\multirow{2}*{Ours$\dag$}  
& Syn.  & ---   & ---   & --- & ---   & ---  & ---  & --- & --- & ---   \\
& Real  & 30.37   & 0.944   & 0.075 & 29.37   & 0.938   & 0.078  & --- & --- & ---   \\

\hline
\multirow{2}*{Ours} 
& Syn.  &\cellcolor{best}39.07 &\cellcolor{best}0.973 &\cellcolor{best}0.026 &\cellcolor{best}37.53 &\cellcolor{best}0.966 &\cellcolor{best}0.024 &\cellcolor{second}36.40 &\cellcolor{second}0.936 &\cellcolor{second}0.018 \\
& Real  & \cellcolor{second}31.63   & \cellcolor{second}0.948   & \cellcolor{second}0.069  & \cellcolor{second}31.43   & \cellcolor{second}0.943   & \cellcolor{second}0.069  & --- & --- & ---   \\
\hline
    \end{tabular}
    \begin{tablenotes}
        \footnotesize
        \item[1] The exposures of input views for NeRF-W are randomly selected from all five exposures to learn five appearance vectors for testing.
        \item[2] A  version of  NeRF (as the upper bound of our method) that  is  trained  from  LDR  images  with  consistent exposures or HDR images.
        \item[$\dag$] An ablation study of our method that models the tone-mapping operations of RGB channels with a single MLP.
     \end{tablenotes}
    \end{threeparttable}
    \label{tb:cmp_syn}
    
\end{table*}




\noindent\textbf{Unit exposure loss.}
Our method recovers radiance $\mathbf{e}$ up to an unknown scale factor $\alpha$ (\ie, $\alpha \mathbf{e}$) via the color reconstruction loss. It is equivalent to add a shift $\ln \alpha$ to the independent variable of function $g$, according to \cref{eq:inverse}, as shown in \cref{fig:ablation_unit_d}. 
As a consequence, we need to add an additional constraint to fix the scale factor $\alpha$. Specifically, we fix the value of $g(0)$ to $C_0$, and the unit exposure loss is defined as:
\begin{equation}\small
	\mathcal{L}_{u} = \|g(0) - C_0 \|_2^2.
\end{equation}
The meaning of this constraint is that the pixels with the value $C_0$ are assumed to have a unit exposure. However, $C_0$ is usually unknown in practice. We generally set the $C_0$ as the midway of the pixel value on real-world scenes. 





Finally, our HDR-NeRF is end-to-end optimized using the following loss:
\begin{equation}\small
	\mathcal{L} = \mathcal{L}_c + \lambda_u\mathcal{L}_u,
\end{equation}
where $\lambda_u$ denotes the weight of unit exposure loss.


\begin{figure*}[t]
    \centering
    \includegraphics[width=\textwidth]{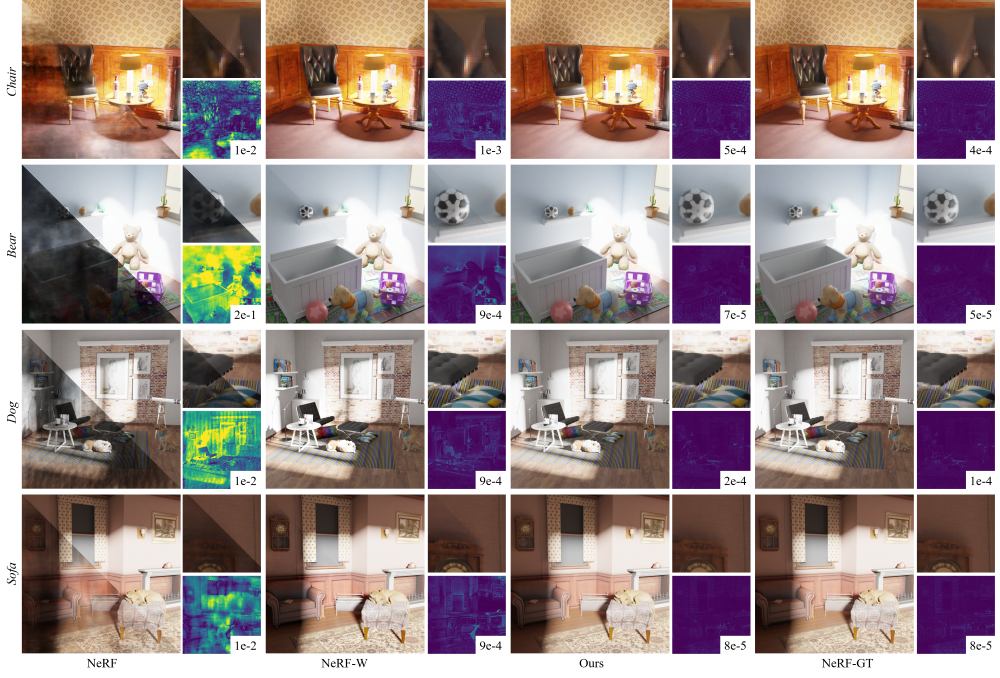}
    \vspace{-5mm}
    \caption{Qualitative comparison of rendered novel LDR view with a novel exposure. The upper triangular images are the ground truth and the lower triangular images are the rendered views. Zoom-in insets and error maps are given on the right. MSE values are on the bottom right of error maps.}
    \label{fig:cmp_LDR_2}
\end{figure*}


\begin{figure*}[!tb]
    \centering
    \subfloat[]{\includegraphics[width=0.162\textwidth]{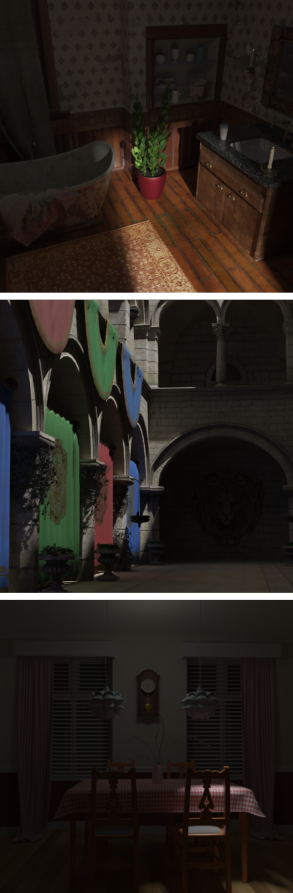}%
    \label{fig:syn_HDR_a}}
    \hfil
    \subfloat[]{\includegraphics[width=0.1615\textwidth]{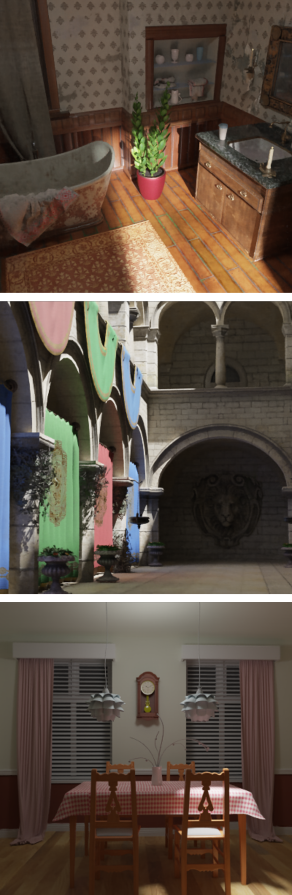}%
    \label{fig:syn_HDR_b}}
    \hfil
    \subfloat[]{\includegraphics[width=0.162\textwidth]{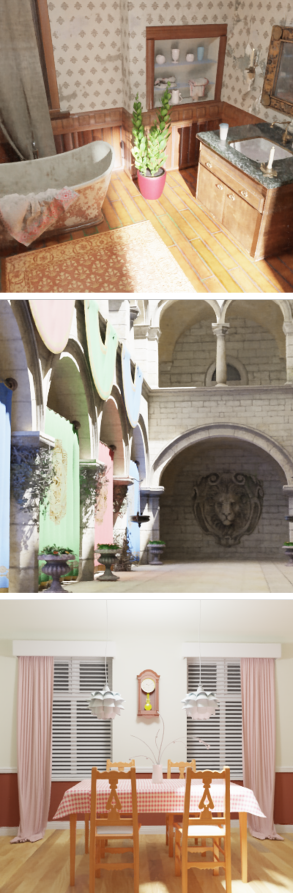}%
    \label{fig:syn_HDR_c}}
    \hfil
    \subfloat[]{\includegraphics[width=0.162\textwidth]{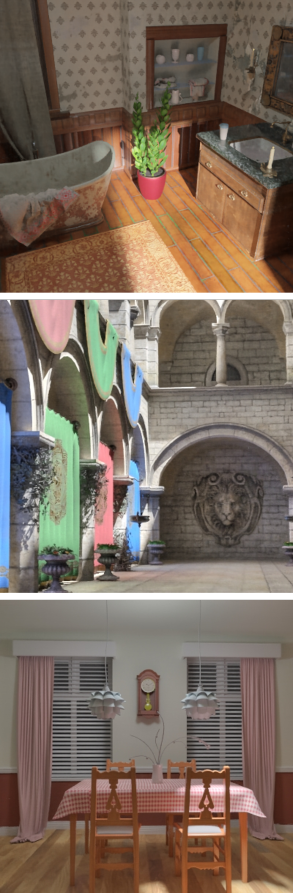}%
    \label{fig:syn_HDR_d}}
    \hfil
    \subfloat[]{\includegraphics[width=0.162\textwidth]{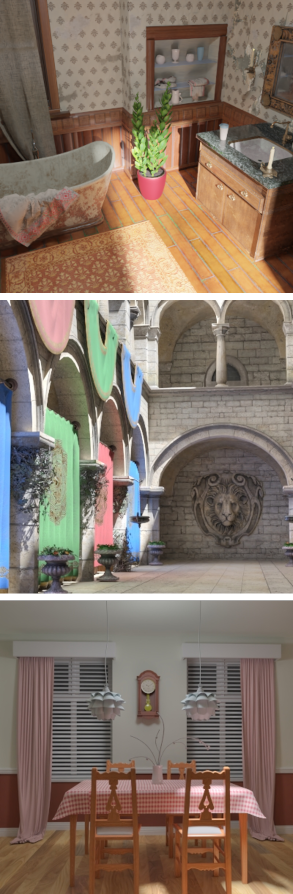}%
    \label{fig:syn_HDR_e}}
    \hfil
    \subfloat[]{\includegraphics[width=0.162\textwidth]{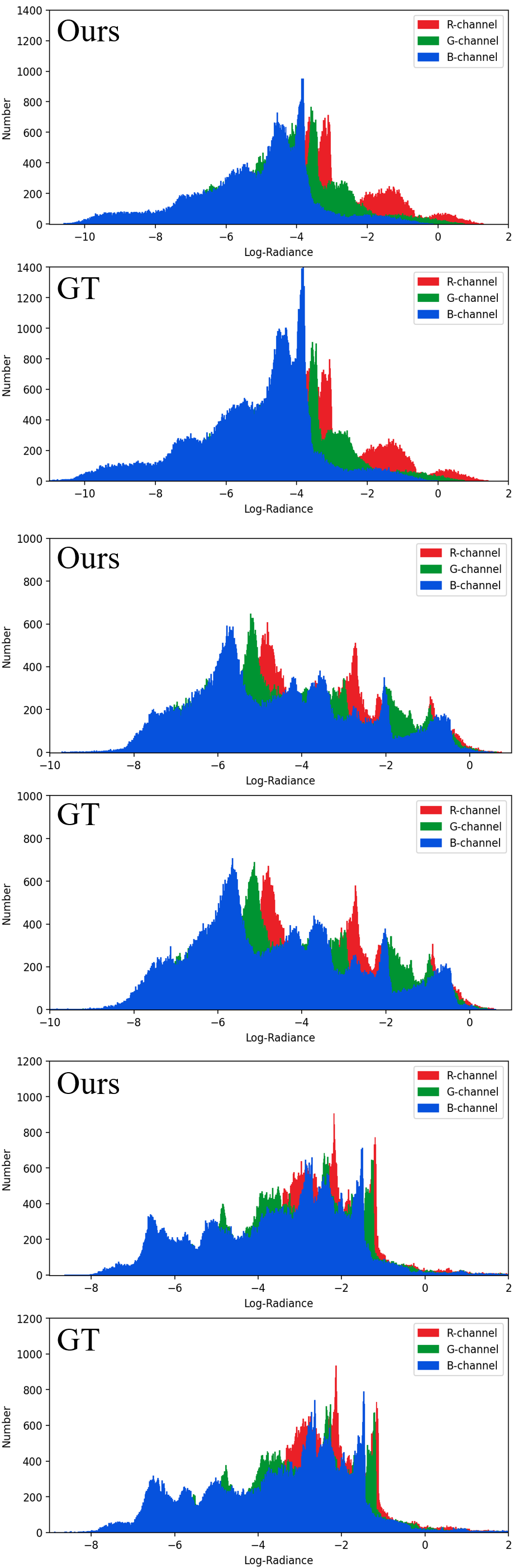}%
    \label{fig:syn_HDR_f}}
    \caption{Qualitative results of our novel LDR views and HDR views on synthetic scenes. (a--c) Our LDR views under different exposures. (d) Our tone-mapped HDR views and (e) ground truth tone-mapped HDR views. (f) Histograms of our novel HDR view (the upper one) and ground truth (the lower one). \bf{Better viewed on screen with zoom in.}}
    \label{fig:syn_HDR}
\end{figure*}

\begin{figure*}[!tb]
    \centering
    \subfloat[\textit{box}]{\includegraphics[width=0.245\textwidth]{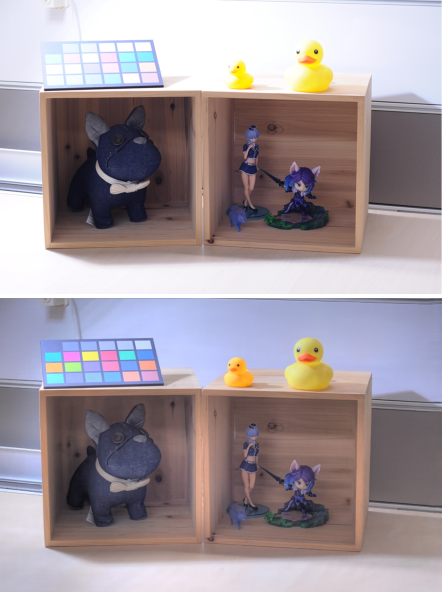}%
    \label{fig:real_HDR_a}}
    \hfil
    \subfloat[\textit{computer}]{\includegraphics[width=0.245\textwidth]{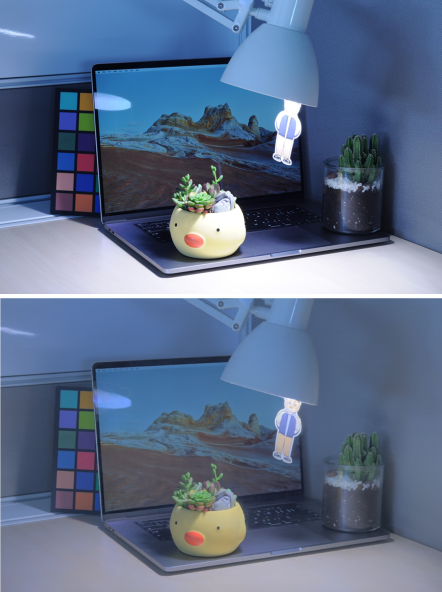}%
    \label{fig:real_HDR_b}}
    \hfil
    \subfloat[\textit{luckycat}]{\includegraphics[width=0.245\textwidth]{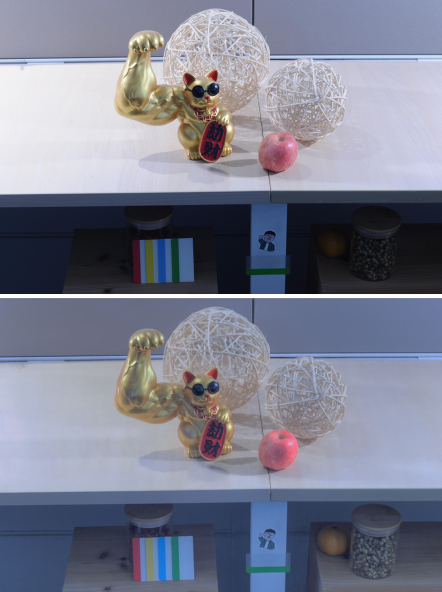}%
    \label{fig:real_HDR_c}}
    \hfil
    \subfloat[\textit{flower}]{\includegraphics[width=0.245\textwidth]{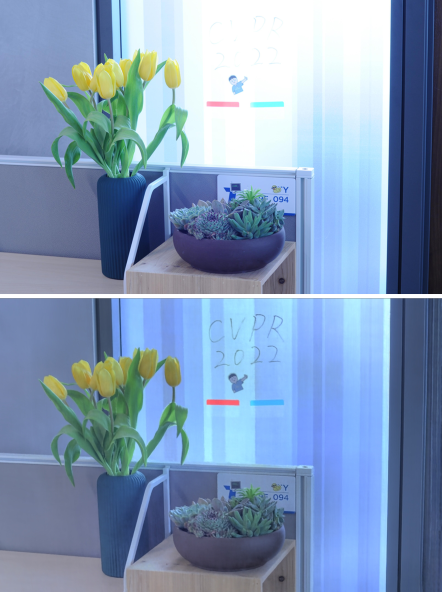}%
    \label{fig:real_HDR_d}}
    \caption{Qualitative results of our novel HDR views on real scenes. Compared with the ground truth LDR views (the first row), our tone-mapped HDR views (the second row) reveal the details of over-exposure and under-exposure areas.}
    \label{fig:real_HDR}
\end{figure*}

\section{Experiments}

\subsection{Implementation Details}
In training and testing phases, an eight-layer MLP with 256 channels is used to predict radiance $\mathbf{e}$ and densities $\sigma$, and three one-layer MLPs with 128 channels to predict RGB values of color $c$ respectively. We sample $64$ points along each ray in the coarse model and $128$ ($64$) points in the fine model on synthetic (real) dataset. The batch size of rays is set to $1024$. As with NeRF, positional encoding \cite{mildenhall2020nerf} is applied for ray origins and ray directions. 
We fix the loss weight $\lambda_u \!\!=\!\! 0.5$ throughout the paper.
The high parameter $C_0$ is $0.5$ on real scenes. To compare with ground truth HDR views, we set $C_0 \!= \!C_{0}^{GT}$ on synthetic scenes, where $C_{0}^{GT}$ denotes the pixel value of ground truth CRF when input logarithm radiance is $0$. We use Adam optimizer \cite{kingma2014adam} (default values $\beta_1=0.9$, $\beta_2=0.999$ and $\epsilon=10^{-7}$) with a learning rate $5\times10^{-4}$ that decays exponentially to $5\times10^{-5}$ over the course of optimization. We optimize a single model for 200K iterations on a single NVIDIA V100 GPU (about one day).

\begin{figure}[!t]
    \centering
    \subfloat[]{\includegraphics[width=0.5\linewidth]{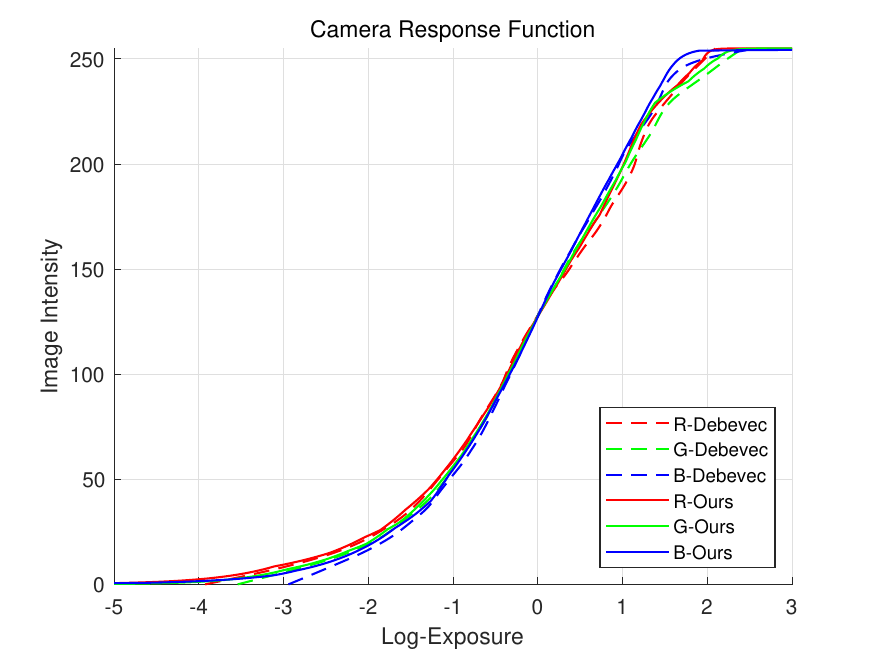}%
    \label{fig:crf_flower_main}}
    \hfil
    \subfloat[]{\includegraphics[width=0.5\linewidth]{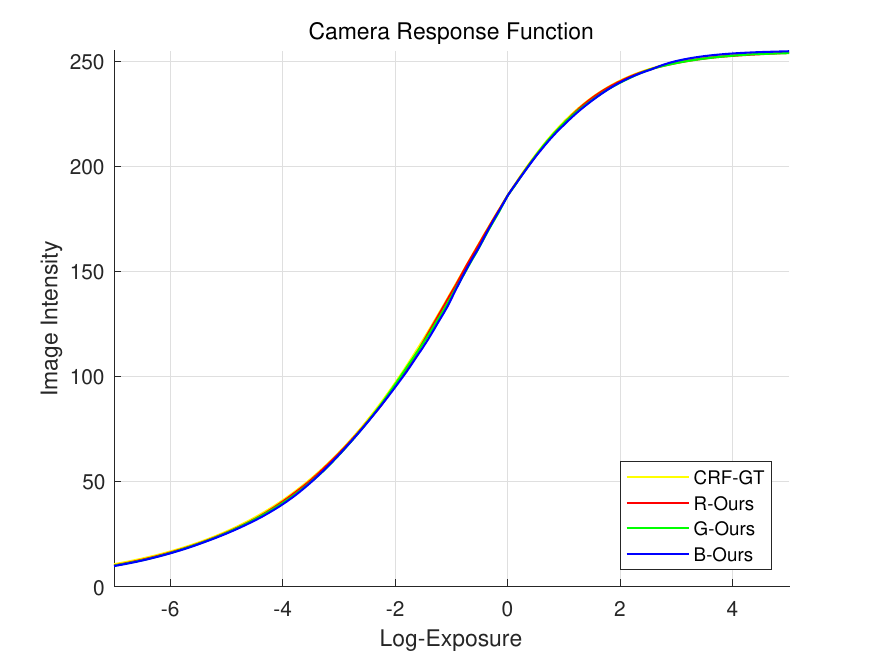}%
    \label{fig:crf_chair_main}}
    \caption{Discrete CRFs estimated by our method on (a) real \textit{flower} scene and (b) synthetic \textit{chair} scene. On the real scene, we calibrate the CRF of digital camera using the method by Debevec and Malik \cite{debevec1997recovering}.}
    \label{fig:crf}
\end{figure}

    

\subsection{Evaluation Dataset and Metrics.}

\noindent\textbf{Dataset.} We evaluate the proposed method on our collected HDR dataset that contains $8$ synthetic scenes rendered with Blender \cite{blender} and $4$ real scenes captured by a digital camera. Images are collected at $35$ different poses in the real dataset, with $5$ different exposure time $\{t_1, t_2, t_3, t_4, t_5\}$ at each pose. For the synthetic dataset, we render $35$ HDR views for each scene and build a tone-mapping function to map these HDR views into LDR images as our inputs (described in supplementary material). The pre-defined tone-mapping function can also be used to evaluate the discrete CRFs estimated by our \textit{tone mapper}.
We select $18$ views with different poses as the training dataset. The exposure time of each input view is randomly selected from $\{t_1, t_3, t_5\}$. $34$ views with exposure time $t_3$ or $t_4$ at the other $17$ poses are chosen as our test dataset. Besides, the HDR views are also used for test. The resolution of each view is $400\times400$ pixels for synthetic scenes and $804\times534$ pixels for real scenes.

\noindent\textbf{Metrics.} We report quantitative performance using PSNR (higher is better) and SSIM (higher is better) metrics, as well as the state-of-the-art LPIPS \cite{zhang2018unreasonable} (lower is better) perceptual metric, which is based on a weighted combination of neural network activations tuned to match human judgments of image similarity \cite{mildenhall2019local}. Since HDR images are usually displayed after a tone-mapping operation, we quantitatively evaluate our HDR views in the tone-mapped domain 
via the $\mu$-law, \ie a simple and canonical operator that is widely used for benchmarking in HDR imaging \cite{kalantari2017deep,yan2019attention,prabhakar2020towards}. The tone-mapping operation is:
\vspace{-2mm}
\begin{equation}\small
	M(E) = \frac{\log (1+\mu E)}{\log (1+\mu)},
	\label{eq:u-tonemap}
\end{equation}
where $\mu$ defines the amount of compression and is always set to $5000$, and $E$ denotes an HDR pixel value which is always scaled to the range $[0, 1]$. To properly show the details in each HDR image for qualitative evaluations, all the HDR results are tone-mapped with Photomatix~\cite{photomatix}.

\subsection{Evaluation}
\noindent\textbf{Baselines.} We compare our method against the following baseline methods. 1) NeRF \cite{mildenhall2020nerf}: the original NeRF method. 2) NeRF-W \cite{martin2021nerf}: unofficial implementation of NeRF in the wild with PyTorch. NeRF-W controls the appearance of rendered views by linearly interpolating their learned appearance vectors, which means that we can not render views by giving the novel exposure time we expect. To facilitate the comparison, the exposure time of input views for NeRF-W are chosen randomly from all the five exposure settings in order to learn five appearance vectors for testing. 3) NeRF-GT (the upper bound of our method): NeRF model trained from LDR views with a consistent exposure or HDR views. 4) Ours$\dag$ (an ablation study): our method that models the tone-mapping operations of RGB channels with a single MLP.

\noindent\textbf{Comparisons.} The quantitative results of rendered novel views on our dataset are shown in \cref{tb:cmp_syn}. Our method outperforms NeRF and NeRF-W on both synthetic and real datasets. Note that only our method can output both LDR and HDR views. Compared with NeRF-GT, our method achieves similar performance for rendering LDR views on the synthetic dataset, while our LDR views have a lower PSNR on real scenes. We notice that our estimated CRF of the blue channel has a bias due to the noise of training views, as seen in \cref{fig:crf_flower_main}, which results in the lower PSNR. As for rendering HDR views, our method is even comparable to NeRF-GT, and we find that directly training the NeRF model from HDR views is hard to produce the expected results, especially on the scene with a larger dynamic range. In addition, we qualitatively compare our method with baselines on rendering novel LDR views with a novel exposure in \cref{fig:cmp_LDR_2}. One can see that the LDR views rendered by our method and NeRF-GT are close to ground truth, but the results of NeRF show serious artifacts because of the varying exposures between input views. The novel views synthesized by NeRF-W appear to be acceptable, yet exhibit inconsistent color with ground truth, as shown in zoom-in insets of \cref{fig:cmp_LDR_2}. Moreover, our novel LDR views with different exposures are shown in \cref{fig:syn_HDR}. It validates that our method can control the exposure of rendered views by giving a specified exposure time.

The novel HDR views are presented in \cref{fig:syn_HDR} and \cref{fig:real_HDR}. It can be seen that the HDR results by our approach (\cref{fig:syn_HDR_d}) are reasonably close to ground truth HDR images (\cref{fig:syn_HDR_e}). Furthermore, compared with LDR views, our tone-mapped HDR views reveal the details of over-exposure and under-exposure areas. We also present the histograms of our and ground truth HDR views in \cref{fig:syn_HDR_f}. The distributions of our histograms are similar to those of ground truth. Besides, discrete CRFs estimated by our method are shown in \cref{fig:crf}, which validates that our \textit{tone mapper} can accurately model the response functions of cameras.

\begin{table}[t]
\small
\centering
\caption{A comparison of our method with $2$ exposures \{$t_1$, $t_5$\}, $3$ exposures \{$t_1$, $t_3$, $t_5$\}, or $5$ exposures: \{$t_1$, $t_2$, $t_3$, $t_4$, $t_5$\}. Metrics (PSNR/SSIM/LPIPS) are averaged over synthetic scenes.}
\begin{tabularx}{\linewidth}{@{}lCCC}
\hline
\multicolumn{1}{l}{} & \multicolumn{1}{c}{LDR-OE} & \multicolumn{1}{c}{LDR-NE} & \multicolumn{1}{c}{HDR} \\
\hline
2  & 32.39/0.954/0.040  & 32.76/0.950/0.036  & 33.00/0.949/0.040 \\
3  & 37.52/0.964/0.022  & 35.73/0.954/0.025  & 37.60/0.963/0.021 \\
5  & 37.73/0.968/0.020  & 36.26/0.960/0.022  & 37.86/0.969/0.019 \\
\hline
\end{tabularx}
\label{tb:num_exp}
\end{table}

\begin{figure}[!t]
    \centering
    \subfloat[with $\mathcal{L}_{u}$]{\includegraphics[width=0.226\linewidth]{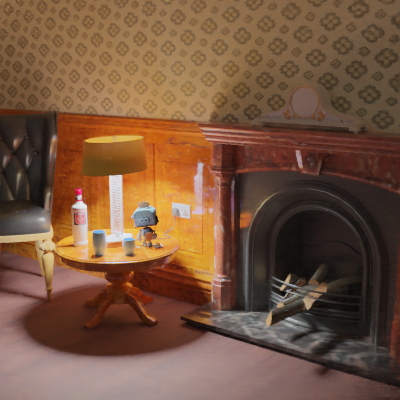}%
    \label{fig:ablation_unit_a}}
    \hfil
    \subfloat[w/o $\mathcal{L}_{u}$]{\includegraphics[width=0.226\linewidth]{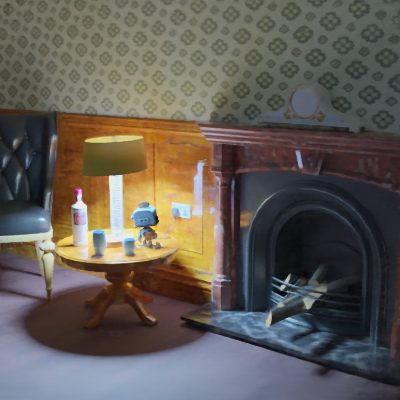}%
    \label{fig:ablation_unit_b}}
    \hfil
    \subfloat[GT]{\includegraphics[width=0.226\linewidth]{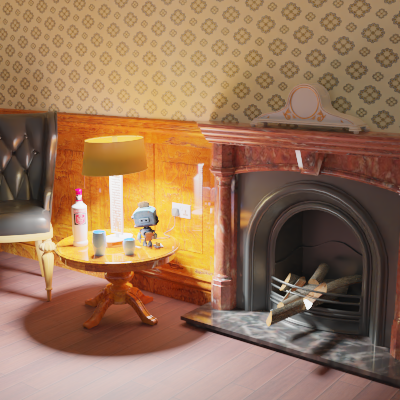}%
    \label{fig:ablation_unit_c}}
    \hfil
    \subfloat[CRFs]{\includegraphics[width=0.29\linewidth]{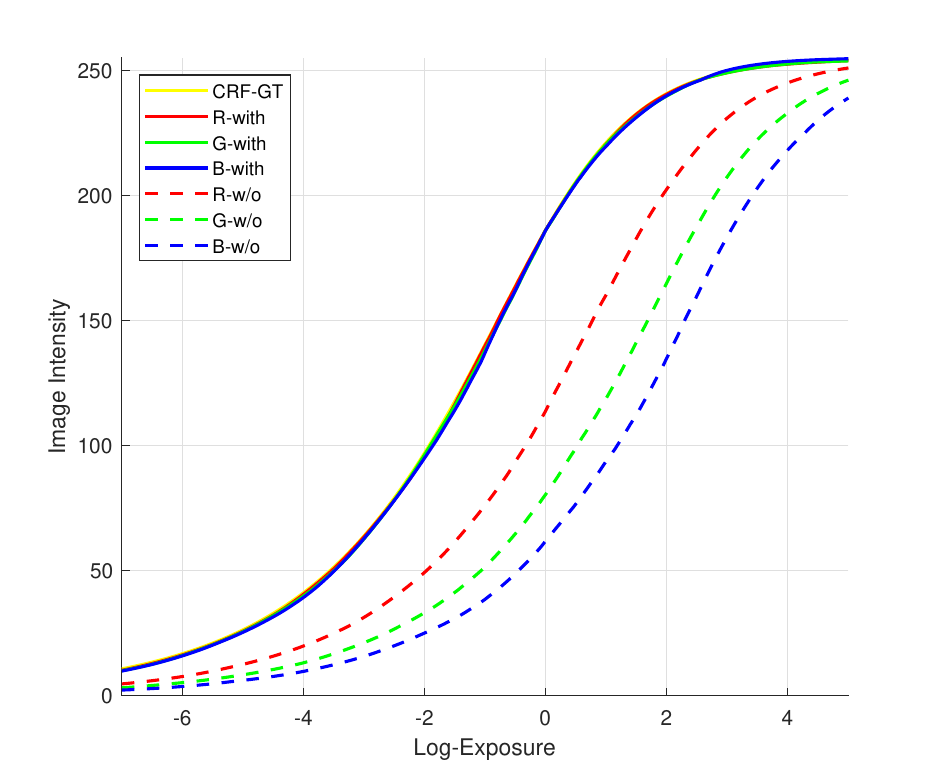}%
    \label{fig:ablation_unit_d}} 
    \caption{Qualitative results with/without unit exposure loss $\mathcal{L}_{u}$. (a--c) The tone-mapped HDR views. (d) The estimated CRFs. \bf{Better viewed on screen with zoom in.}}
    \label{fig:ablation_unit}
\end{figure}

\noindent\textbf{Ablation Studies.} 1) Theoretically, recovering a camera response curve requires a minimum of two exposures \cite{debevec1997recovering}. We investigate the influence of the number of exposures in \cref{tb:num_exp}, where the number is set to $\{2,3,5\}$ respectively. We can see that the performance of the proposed method improves with the number of exposures. The results are close when the number is set to $3$ or $5$, and both significantly outperform the results of $2$ exposures. Thereby, using $3$ exposures is a reasonable choice. 2) The ablation study of unit exposure loss $\mathcal{L}_{u}$ is presented in \cref{tb:unit_loss} and \cref{fig:ablation_unit}. \Cref{tb:unit_loss} shows that our method produces better quantitative results with the unit exposure loss, especially on rendering HDR views. The HDR images rendered by the approach without unit exposure loss suffer from severe chromatic aberration (\cref{fig:ablation_unit_b}) due to the different shifts of three estimated CRF curves (\cref{fig:ablation_unit_d}). 3) Since the RGB channels have the same CRF in synthetic scenes, we evaluate the efficiency of modeling the CRF with three MLPs on real scenes. As shown in \cref{tb:cmp_syn}, when three channels are processed independently, our method achieves superior results.

\begin{table}[!tb]
\small
\centering
\caption{Quantitative results with/without unit exposure loss $\mathcal{L}_{u}$. Metrics are averaged over synthetic scenes.}
\begin{tabularx}{\linewidth}{@{}lCCCCCC}
\hline
       & \multicolumn{3}{c}{with $\mathcal{L}_{u}$} & \multicolumn{3}{c}{w/o $\mathcal{L}_{u}$} \\
\cmidrule(lr){2-4} \cmidrule(lr){5-7}
       & PSNR$\uparrow$  & SSIM$\uparrow$ & LPIPS$\downarrow$  & PSNR$\uparrow$  & SSIM$\uparrow$ & LPIPS$\downarrow$ \\
\hline
LDR-OE & 37.52  & 0.964  & 0.022  & 36.48  & 0.957  & 0.030 \\
LDR-NE & 35.73  & 0.954  & 0.025  & 34.77  & 0.947  & 0.035 \\
HDR    & 37.60  & 0.963  & 0.021  & 13.35  & 0.765  & 0.163 \\
\hline
\end{tabularx}
\label{tb:unit_loss}
\end{table}

\noindent\textbf{Limitations.} Recovering an HDR radiance field from a series of LDR images with different exposures is challenging. Similar to the classic HDR radiance map recovering method \cite{debevec1997recovering}, our recovered HDR radiance field is relative. There are three unknown scaling factors (for RGB channels) that relate the recovered radiance to absolute radiance. Consequently, different choices of these factors will recover HDR radiance fields with different white balances. Besides, our \textit{tone mapper} models the camera coarsely without considering the effect of ISO gain and aperture for exposures. %

\section{Conclusion}
We have proposed a novel method to recover the high dynamic range neural radiance field from a set of LDR views with different exposures. Our method not only renders novel HDR views without ground-truth HDR supervision, but also produces high-fidelity LDR views with specified exposures. The core of the method is modeling the process that captures scene radiance and maps them into pixel values. Compared with prior works, our method performs better in rendering LDR views. Importantly, to our knowledge our method is the first neural rendering method that synthesizes novel views with high dynamic range. Code and models will be made available to the research community to facilitate reproducible research.

\noindent\textbf{Acknowledgements.} The work was supported by NSFC under Grant 62031023. The authors thank Li Ma and Xiaoyu Li for their instructive and useful advice. 

{\small
\bibliographystyle{ieee_fullname}
\bibliography{egbib}
}

\clearpage
\section*{Supplemental Materials}
\setcounter{section}{0}
\renewcommand\thesection{\Alph{section}}
\section{Overview}
The supplementary material shows the additional implementation details of our method, baselines, and our collected HDR dataset. Additional results are also presented to further demonstrate the superior performance of our method. We \textbf{strongly} encourage the reader to see our video supplementary, in which we present our results on the test scenes and comparisons with baselines.

\section{Additional Implementation Details}
Our code is built upon the PyTorch implementation of NeRF (\href{https://github.com/yenchenlin/nerf-pytorch}{https://github.com/yenchenlin/nerf-pytorch}). During the training and testing, the rays are mapped from camera space to the normalized device coordinate (NDC) space \cite{mildenhall2020nerf}. The inference code and one model are provided in our supplementary materials. 

To evaluate our estimated CRFs on synthetic scenes, we build a simple global tone-mapping function based on the classical Reinhard tone-mapping \cite{reinhard2002photographic}. Using this function, the HDR views rendered by Blender are tone-mapped into LDR views. We then take the LDR views as our inputs.  The simple tone-mapping function is defined as: 
\begin{equation}\small
	M(E) = {\left(\frac{E}{E+1}\right)}^{\frac{1}{2.2}},
	\label{eq:simple-tonemap}
\end{equation}
where $E$ is the HDR pixel value. To generate LDR views with different exposure, we use the \textit{exposure value} $EV$ to scale the HDR pixel value $E$ (\ie $2^{EV}E$). We introduce the exposure value $EV$ into \cref{eq:simple-tonemap}:
\begin{equation}\small
	M(E, EV) = {\left(\frac{2^{EV} E}{2^{EV} E+1}\right)}^{\frac{1}{2.2}},
	\label{eq:simple-tonemap-ev}
\end{equation}
where $2^{EV}$ is also can be considered as the exposure time in our paper, that's $\Delta t = 2^{EV}$. 


\section{Baseline Methods Implementation Details}
The import parameters of baseline methods, such as number of samples per ray, position encoding, and batch size, are all set as same as these of us for a fair comparison. All the models are trained with Adam about $200,000$ iterations.

\noindent\textbf{NeRF}: We use the PyTorch implementation of NeRF code open-source at  \href{https://github.com/yenchenlin/nerf-pytorch}{https://github.com/yenchenlin/nerf-pytorch}.

\noindent\textbf{NeRF-W}: The code of NeRF-W is provided at   \href{https://github.com/kwea123/nerf\_pl/tree/nerfw}{https://github.com/kwea123/nerf\_pl/tree/nerfw}, which is an unofficial implementation of NeRF-W using PyTorch (PyTorch-lightning). 

\noindent\textbf{NeRF-GT}: The NeRF-GT is a version of NeRF that is directly trained from LDR views with consistent exposures or HDR views, which can be considered as the upper bound of our method. When we train the NeRF model from HDR views, the predicted HDR pixel values are tone-mapped into LDR pixel values and then compared to the tone-mapped ground truth. However, we find that it is difficult to ensure all the areas of a scene are encoded well by NeRF model, due to the high dynamic range of the scenes, even though we use the tone-mapped predicted color to calculate the loss.

\section{HDR Dataset Details}
Since no dataset is appropriate for the task of novel HDR views synthesis, we collect a new dataset for the evaluation of our method. Most 3D models used in our dataset are provided at \href{https://sketchfab.com/feed}{https://sketchfab.com/feed}.
All the licenses of 3D models will be attached, when we release our dataset. The HDR views for each scene are rendered with Blender's Cycles path-tracer \cite{blender}. For real-world scenes, the LDR views with different exposures are captured by a Nikon D90 camera. We set the ISO gain to 200 and aperture to $f/6.7$. We calibrate a set of LDR images using an open-source software package COLMAP \cite{schonberger2016structure}. The calibration setting of COLMAP follows the one of LLFF \cite{mildenhall2019local}. We also capture 10 images with different exposures for each scene to calibrate the CRF of the Nikon D90 camera. The CRFs are calibrated with the classical method by Debevec and Malik \cite{debevec1997recovering}.

\begin{figure}[t]
  \centering
  \includegraphics[width=\linewidth]{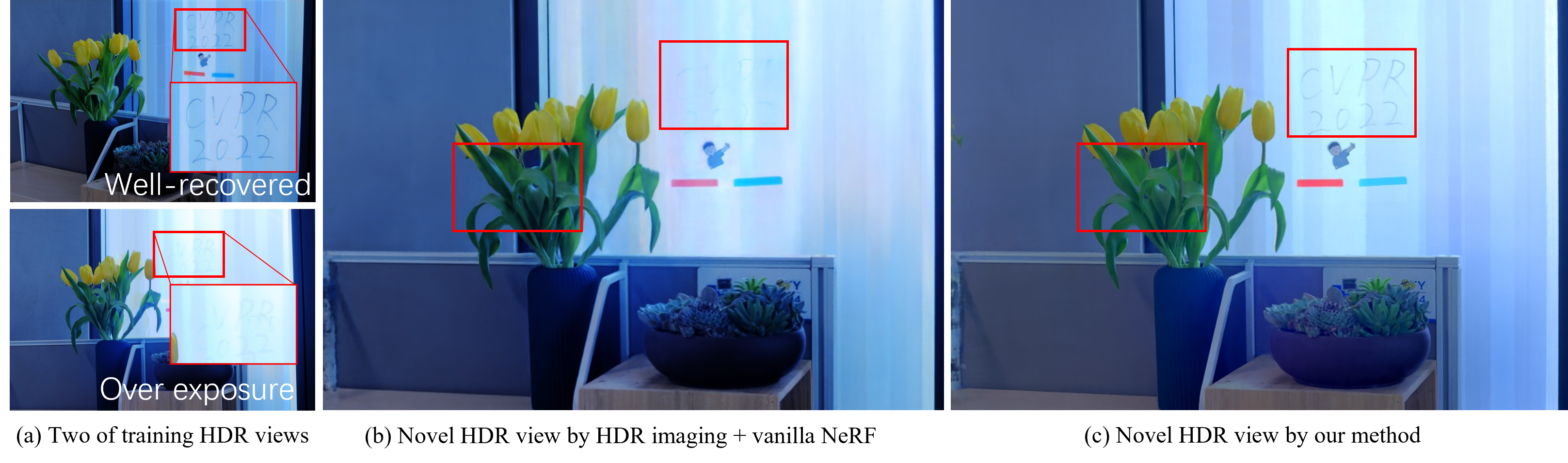}
   \caption{The comparisons with HDR imaging $+$ vanilla NeRF. All the HDR images are tone-mapped with same hyperparameters.}
   \label{fig:1}
\end{figure}

\section{Additional Results}
\begin{figure*}[t]
    \centering
    \includegraphics[width=\textwidth]{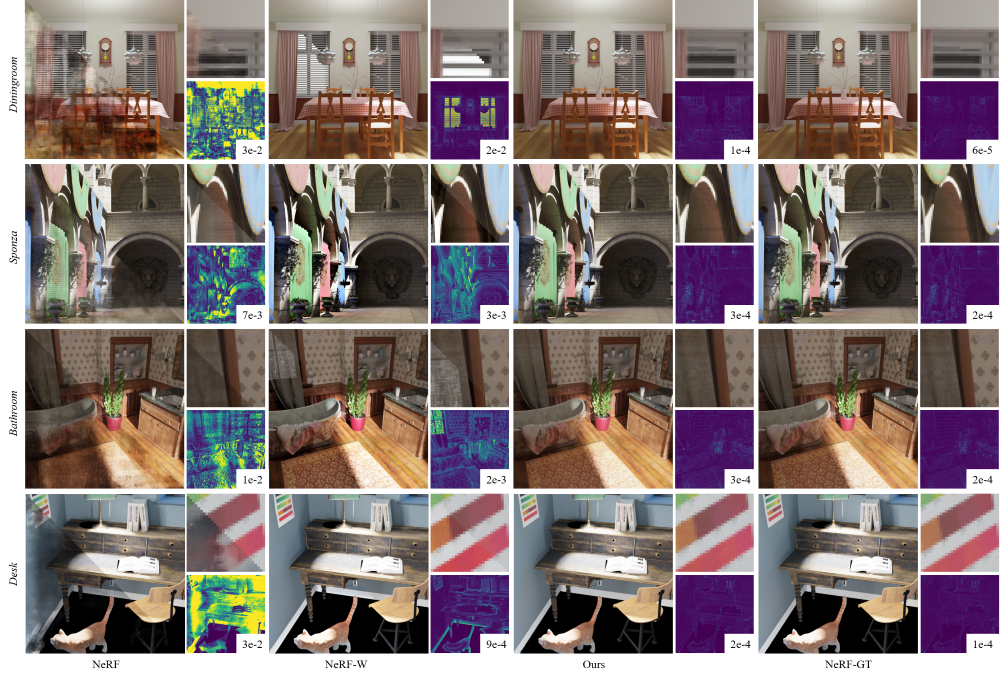}
    \vspace{-5mm}
    \caption{Qualitative comparisons of novel LDR views with novel exposures. The upper triangular images are the ground truth and the lower triangular images are the rendered views. Zoom-in insets and error maps are given on the right. MSE values are on the bottom right of error maps.}
    \label{fig:cmp_LDR_1}
\end{figure*}

\begin{figure*}[!tb]
    \centering
    \includegraphics[width=\textwidth]{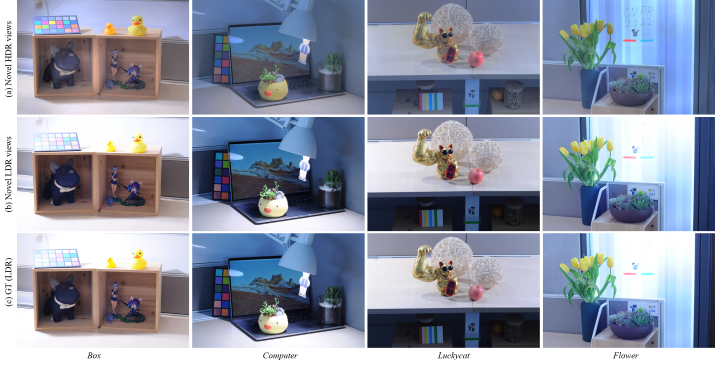}
    \vspace{-5mm}
    \caption{Qualitative results of our novel views on real scenes. (a) Our tone-mapped HDR views using Photomatix \cite{photomatix}. (b) Our novel LDR views with novel exposures. (c) Ground truth LDR views. }
    \label{fig:cmp_real_all}
\end{figure*}

\begin{figure*}[!tb]
    \centering
    \includegraphics[width=\textwidth]{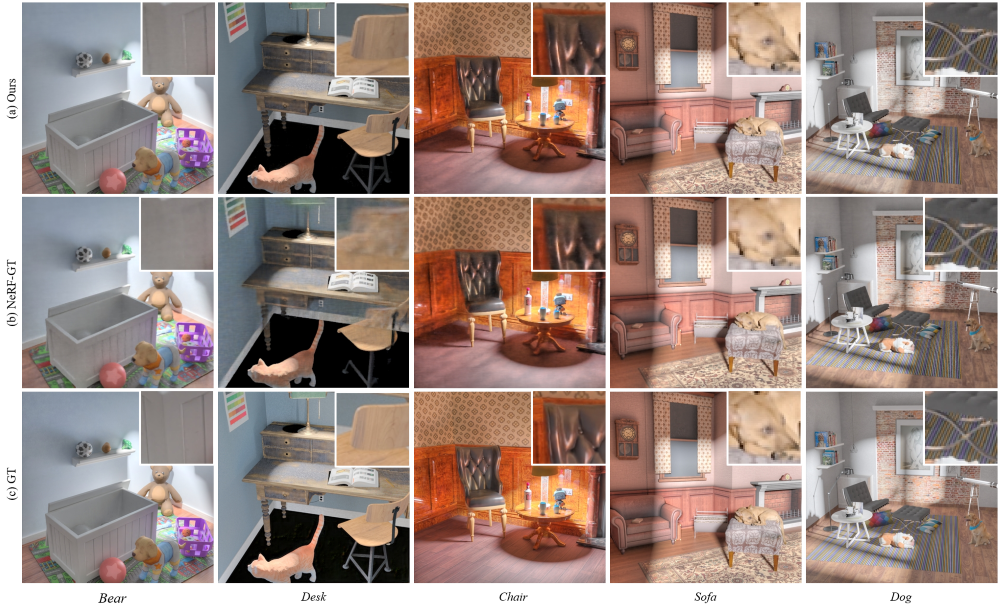}
    \vspace{-5mm}
    \caption{Qualitative results of our novel HDR views on synthetic scenes. All the HDR views are tone-mapped using Photomatix \cite{photomatix}. (a) Our novel HDR views. (b) The novel HDR views by NeRF-GT that a NeRF model is tarined from HDR views. (c) The ground truth HDR views.}
    \label{fig:cmp_syn_all}
\end{figure*}

\begin{figure*}[!t]
    \centering
    \subfloat[\textit{Bathroom}]{\includegraphics[width=0.25\textwidth]{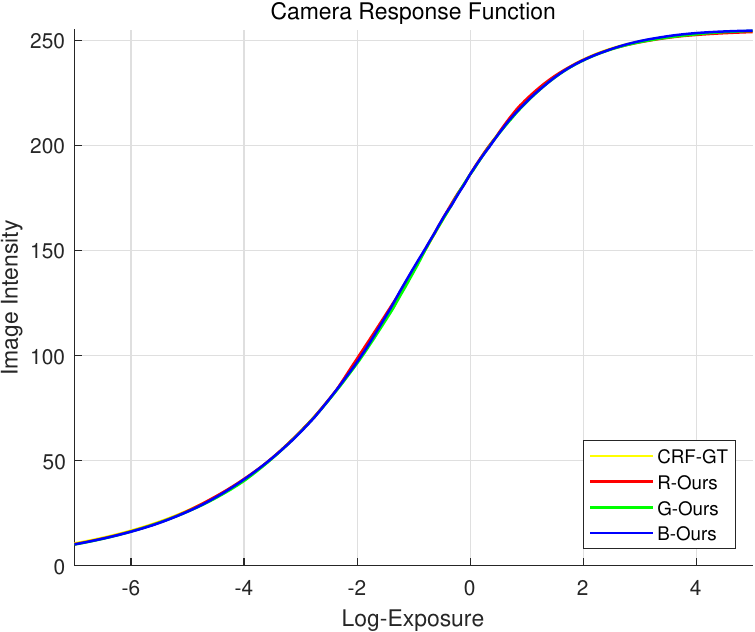}%
    \label{fig:crf_bathroom}}
    \hfil
    \subfloat[\textit{Bear}]{\includegraphics[width=0.25\textwidth]{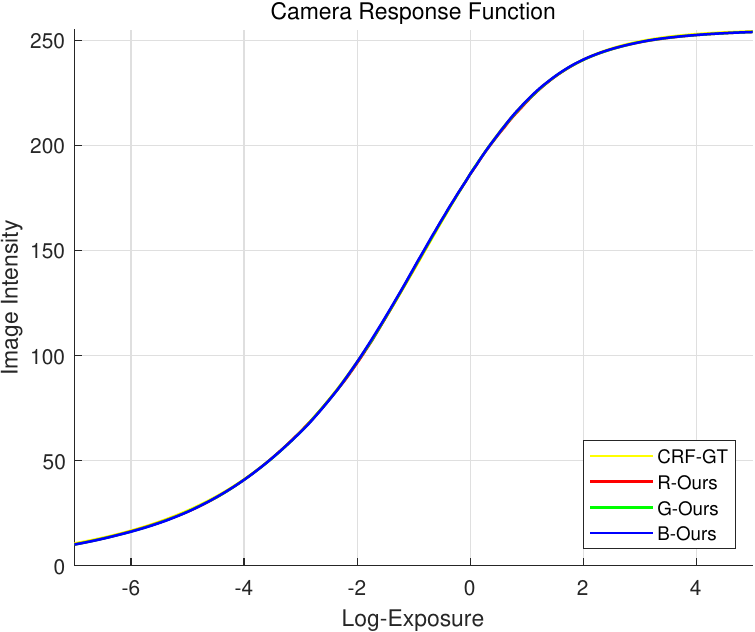}%
    \label{fig:crf_bear}}
    \hfil
    \subfloat[\textit{Chair}]{\includegraphics[width=0.25\textwidth]{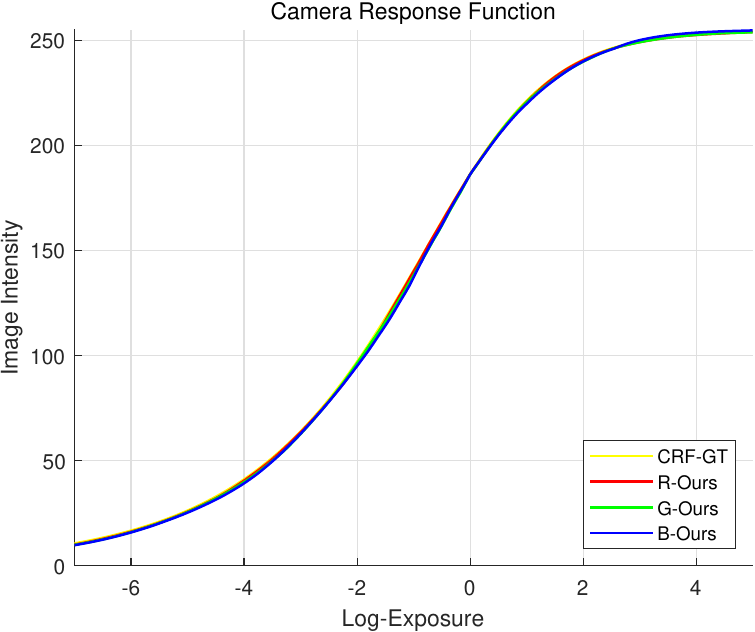}%
    \label{fig:crf_chair}}
    \hfil
    \subfloat[\textit{Desk}]{\includegraphics[width=0.25\textwidth]{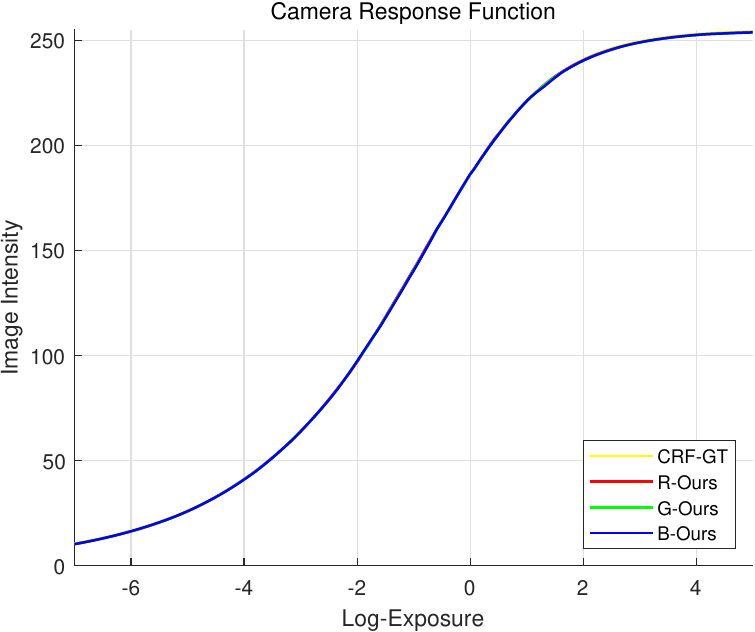}%
    \label{fig:crf_desk}}
    \\
    \subfloat[\textit{Diningroom}]{\includegraphics[width=0.25\textwidth]{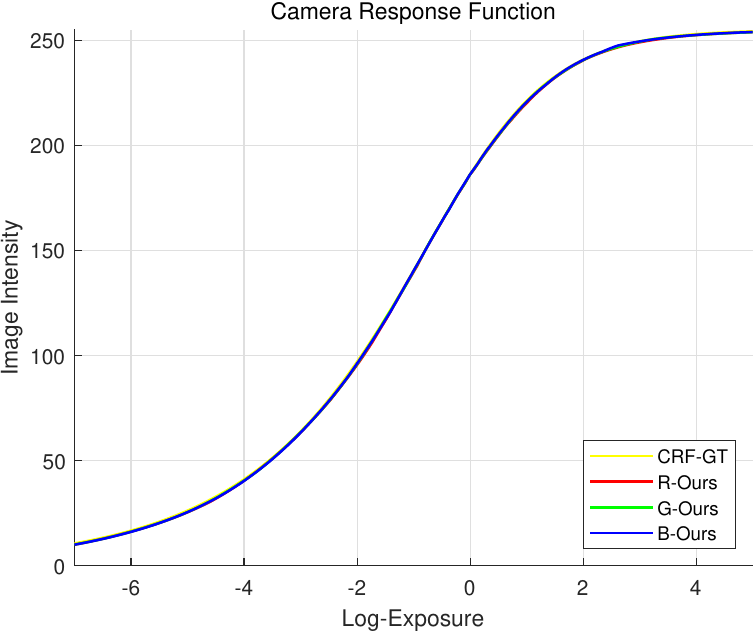}%
    \label{fig:crf_diningroom}}
    \hfil
    \subfloat[\textit{Dog}]{\includegraphics[width=0.25\textwidth]{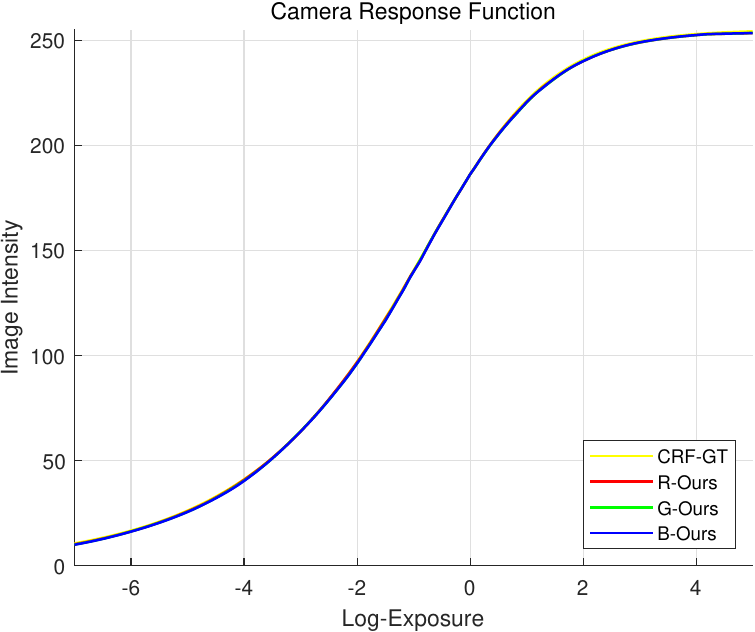}%
    \label{fig:crf_dogroom}}
    \hfil
    \subfloat[\textit{Sofa}]{\includegraphics[width=0.25\textwidth]{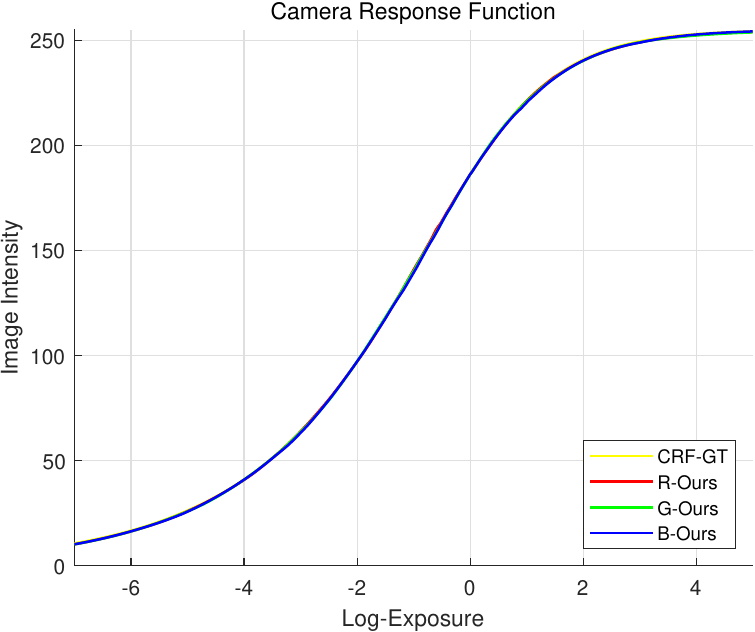}%
    \label{fig:crf_sofa}}
    \hfil
    \subfloat[\textit{Sponza}]{\includegraphics[width=0.25\textwidth]{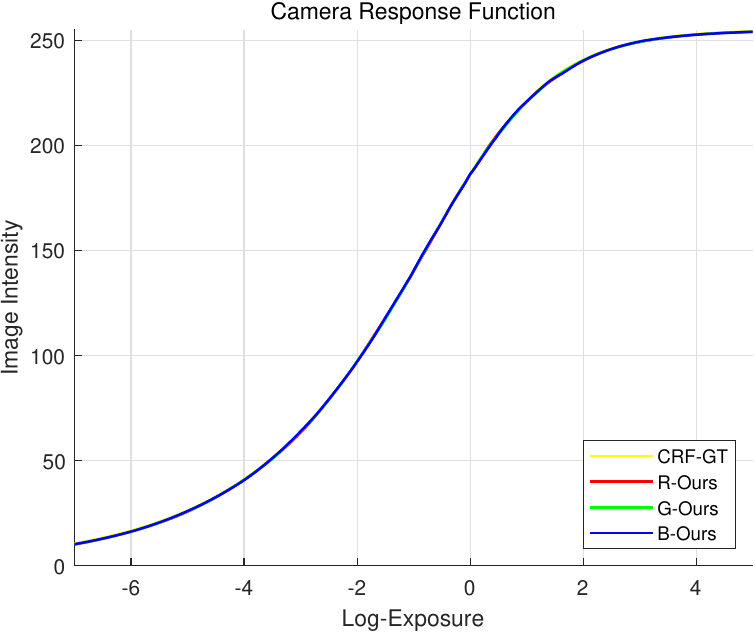}%
    \label{fig:crf_sponza}}
    \\
    \subfloat[\textit{Box}]{\includegraphics[width=0.25\textwidth]{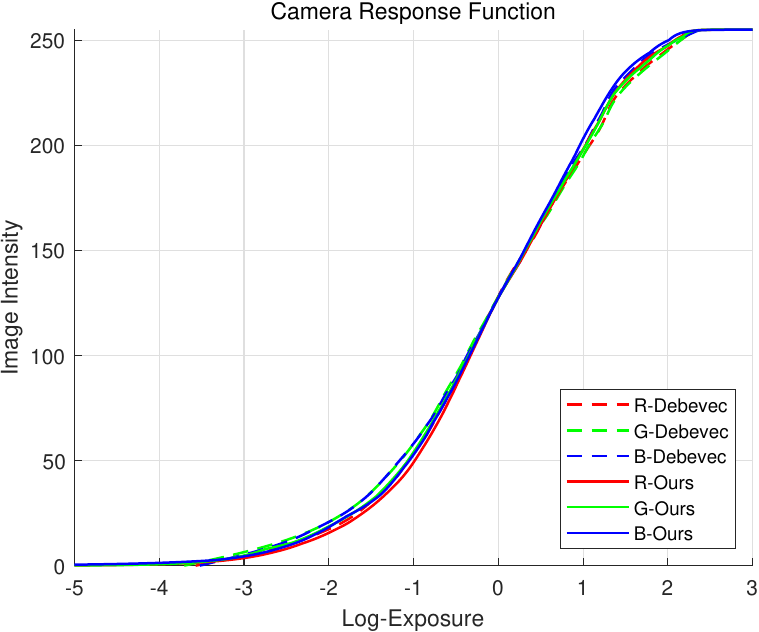}%
    \label{fig:crf_colorboard}}
    \hfil
    \subfloat[\textit{Flower}]{\includegraphics[width=0.25\textwidth]{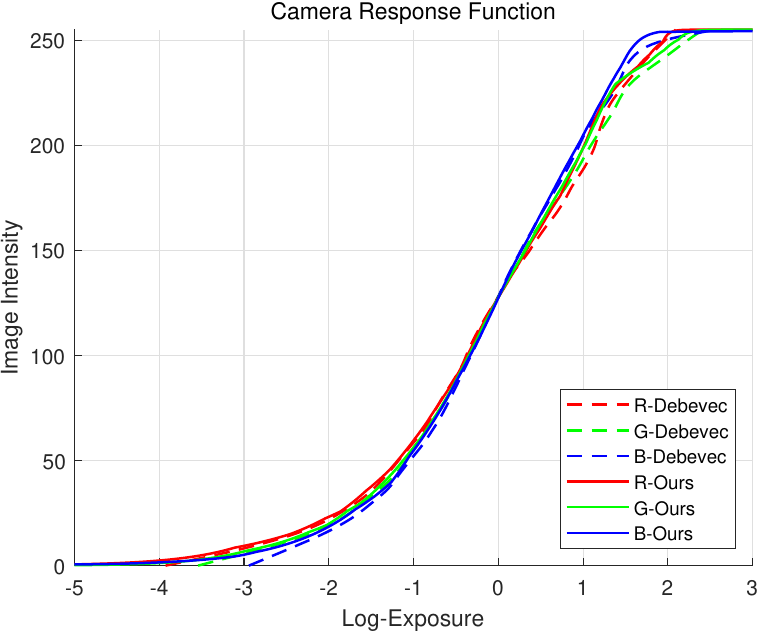}%
    \label{fig:crf_flower}}
    \hfil
    \subfloat[\textit{Computer}]{\includegraphics[width=0.25\textwidth]{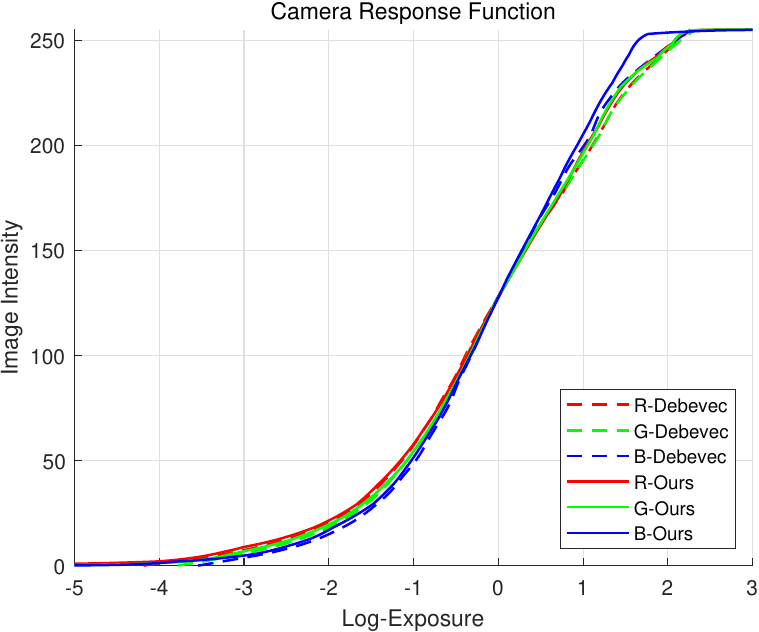}%
    \label{fig:crf_computer}}
    \hfil
    \subfloat[\textit{Luckycat}]{\includegraphics[width=0.25\textwidth]{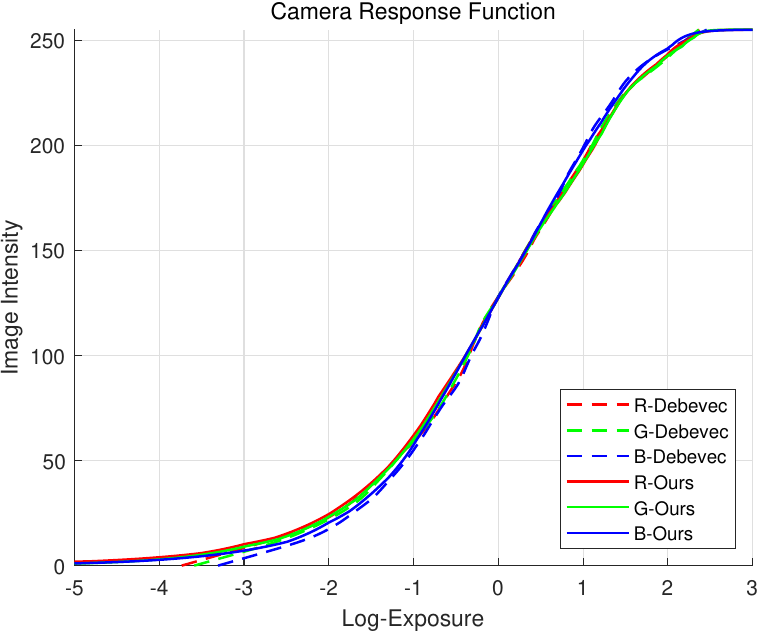}%
    \label{fig:crf_luckycat}}
    
    \caption{All the discrete CRFs estimated by our method on (a--h) synthetic scenes and (i--l) real-world scenes. On real-world scenes, we calibrate the CRF of digital camera using the method by Debevec and Malik \cite{debevec1997recovering}.}
    \label{fig:crf_all}
\end{figure*}

The additional per-scene comparisons with baseline methods on synthetic scenes are shown in \cref{tb:cmp_syn1} and \cref{tb:cmp_syn2}. \Cref{tb:cmp_real} includes a breakdown of the quantitative results on real scenes presented in the main paper into per-scene metrics. The quantitative results further validate that our method outperforms the baseline methods. Figures \ref{fig:cmp_LDR_1}, \ref{fig:cmp_real_all} and \ref{fig:cmp_syn_all} show the qualitative results of our method and baselines. It can be seen that our method can accurately control the exposure of rendered LDR views compared NeRF-W, and the results by our method are reasonable close to those of NeRF-GT (the upper bound). On the other hand, our method can better reconstruct the small textures on rendering HDR views, as shown in \cref{fig:cmp_syn_all}. Finally, all the CRFs estimated by our method are exhibited in \cref{fig:crf_all}, which demonstrates that our method correctly  models the tone-mapping operation of the camera. We have also tried to concatenate the $\ln e$ and $ \ln \Delta t$ then feed them into the tone-mapper. Our method produces similar LDR and HDR results (PSNR: $ \pm 0.1$, SSIM:$ \pm 0.02$, LPIPS:$ \pm 0.01$).

We have tried to reconstruct HDR views using an HDR imaging method \cite{wu2018deep} then train vanilla NeRF, where LDR views ($\{t_1, t_3, t_5\}$) with small disparity are used to reconstruct HDR views. Some results reconstructed by the image-wise HDR imaging method are view inconsistent (\cref{fig:1} (a)) since the radiance scale of each view is different, which leads NeRF or IBR to render the novel views with artifacts (\cref{fig:1} (b)). HDR imaging $\!+\!$ vanilla NeRF is straightforward. The HDR views captured by off-the-shelf cameras are also view-dependent and the radiance scales vary with the poses. Therefore, we propose a novel method for recovering radiance fields from LDR views. Compared with HDR imaging $\!+\!$ vanilla NeRF, our method is an \textbf{end-to-end framework with fewer inputs and better performance.} The auto-exposure scenes can also be handled by modeling more camera settings, such as ISO and aperture. Our method can recover the radiance field and render novel HDR views from an auto-exposure video. Moreover, the exposure can also be learned, just like appearance vectors in NeRF-W.

\begin{table*}[!th]
    \small
    \centering
    \caption{Quantitative comparisons with baseline methods on four synthetic scenes. LDR-OE denotes the average LDR results with exposure $t_1$, $t_3$, and $t_5$. LDR-NE denotes the average LDR results with exposure $t_2$, and $t_4$. HDR denotes the HDR results. We color code each column as \colorbox{best}{best} and \colorbox{second}{second best}.}
    
    \begin{threeparttable}
    \begin{tabular}{@{}p{49pt}|p{39pt}|p{20pt}<{\centering} p{20pt}<{\centering} p{20pt}<{\centering} p{20pt}<{\centering} p{20pt}<{\centering} p{20pt}<{\centering} p{20pt}<{\centering} p{20pt}<{\centering} p{20pt}<{\centering} p{20pt}<{\centering} p{20pt}<{\centering} p{20pt}<{\centering}}
        \hline 
        & & \multicolumn{3}{c}{\textit{Diningroom}} 
        & \multicolumn{3}{c}{\textit{Sponza}}
        & \multicolumn{3}{c}{\textit{Bathroom}}
        & \multicolumn{3}{c}{\textit{Desk}} \\
        \cline{3-14}
    &   & PSNR$\uparrow$ & SSIM$\uparrow$ & LPIPS$\downarrow$ & PSNR$\uparrow$ & SSIM$\uparrow$ & LPIPS$\downarrow$ & PSNR$\uparrow$ & SSIM$\uparrow$ & LPIPS$\downarrow$ & PSNR$\uparrow$ & SSIM$\uparrow$ & LPIPS$\downarrow$\\
        \hline

\multirow{3}*{NeRF\cite{mildenhall2020nerf}} 
& LDR-OE  & 12.50    & 0.378    & 0.600    & 16.39   & 0.664   & 0.219  & 14.59    & 0.429   & 0.424   & 15.29  & 0.645  & 0.249  \\
& LDR-NE  &  ---    & ---   &  ---    & ---     &  ---      & ---      &  ---      & ---   &  ---      & ---   &   ---   & ---       \\
& HDR &  ---  & ---  &  ---      & ---     &  ---      & ---      &  ---      & ---   &  ---      & ---   &   ---   & ---      \\
\hline

\multirow{3}*{NeRF-W\tnote{1} \cite{martin2021nerf}} 
& LDR-OE  & 32.25    & 0.979    & 0.016    & 24.50   & 0.908   & 0.037  & 29.64    & 0.900   & 0.055   & 30.21  & 0.958  & 0.030  \\
& LDR-NE   & 32.53    & 0.972    & 0.019    & 24.32   & 0.904   & 0.042  & 26.98    & 0.881   & 0.066   & 29.60  & 0.950  & 0.034  \\
& HDR &  ---    & ---   &  ---    & ---     &  ---      & ---      &  ---      & ---   &  ---      & ---   &   ---   & ---   \\
\hline

\multirow{3}*{Ours} 
& LDR-OE & \cellcolor{second}41.23    & \cellcolor{second}0.986    & \cellcolor{second}0.010
         & \cellcolor{second}34.49   & \cellcolor{second}0.958   & \cellcolor{second}0.034  
         & \cellcolor{second}36.26    & \cellcolor{second}0.949   & \cellcolor{second}0.037   
         & \cellcolor{second}37.84  & \cellcolor{second}0.972  & \cellcolor{second}0.023  \\
         
& LDR-NE & \cellcolor{second}37.99    & \cellcolor{second}0.979    & \cellcolor{second}0.013
         & \cellcolor{second}33.41   & \cellcolor{second}0.950   & \cellcolor{second}0.038  
         & \cellcolor{second}33.44    & \cellcolor{second}0.926   & \cellcolor{second}0.046   
         & \cellcolor{second}35.26  & \cellcolor{second}0.960  & \cellcolor{second}0.029  \\
         
& HDR    & \cellcolor{second}38.57    & \cellcolor{second}0.981    & \cellcolor{second}0.015    
         & \cellcolor{second}32.33   & \cellcolor{best}0.939   & \cellcolor{second}0.049  
         & \cellcolor{best}33.97    & \cellcolor{best}0.925   & \cellcolor{best}0.048   
         &  \cellcolor{best}43.38  &  \cellcolor{best}0.993  &  \cellcolor{best}0.007  \\
\hline

\multirow{3}*{NeRF-GT\tnote{2} \cite{mildenhall2020nerf}} 
& LDR-OE  & \cellcolor{best}43.66    & \cellcolor{best}0.991    & \cellcolor{best}0.007   
          & \cellcolor{best}37.25   & \cellcolor{best}0.973   & \cellcolor{best}0.020  
          & \cellcolor{best}38.51    & \cellcolor{best}0.964   & \cellcolor{best}0.027   
          & \cellcolor{best}39.22  & \cellcolor{best}0.978  & \cellcolor{best}0.017  \\

& LDR-NE  & \cellcolor{best}41.14    & \cellcolor{best}0.989    & \cellcolor{best}0.007    
          & \cellcolor{best}34.55   & \cellcolor{best}0.958   & \cellcolor{best}0.031  
          & \cellcolor{best}35.42    & \cellcolor{best}0.949   & \cellcolor{best}0.030   
          & \cellcolor{best}37.46  & \cellcolor{best}0.973  & \cellcolor{best}0.020  \\
          
& HDR     & \cellcolor{best}42.49    & \cellcolor{best}0.989    & \cellcolor{best}0.002    
          & \cellcolor{best}32.66   & \cellcolor{second}0.913   & \cellcolor{best}0.012  
          & \cellcolor{second}30.72    & \cellcolor{second}0.798   & \cellcolor{second}0.039   
          & \cellcolor{second}41.15  & \cellcolor{second}0.975  & \cellcolor{second}0.015  \\

        \hline
    \end{tabular}
    \begin{tablenotes}
        \footnotesize
        \item[1] The exposures of input views for NeRF-W are randomly selected from all five exposures to learn five appearance vectors for testing.
        \item[2] A  version of  NeRF (as the upper bound of our method) that  is  trained  from  LDR  images  with  consistent exposures or HDR images.
     \end{tablenotes}
    \end{threeparttable}
    \label{tb:cmp_syn1}
    
\end{table*}

\begin{table*}[th]
    \small
    \centering
    \caption{Quantitative comparisons with baseline methods on four synthetic scenes. LDR-OE denotes the average LDR results with exposure $t_1$, $t_3$, and $t_5$. LDR-NE denotes the average LDR results with exposure $t_2$, and $t_4$. HDR denotes the HDR results. We color code each column as \colorbox{best}{best} and \colorbox{second}{second best}.}
    
    \begin{threeparttable}
    \begin{tabular}{@{}p{49pt}|p{39pt}|p{20pt}<{\centering} p{20pt}<{\centering} p{20pt}<{\centering} p{20pt}<{\centering} p{20pt}<{\centering} p{20pt}<{\centering} p{20pt}<{\centering} p{20pt}<{\centering} p{20pt}<{\centering} p{20pt}<{\centering} p{20pt}<{\centering} p{20pt}<{\centering}}
        \hline 
        & & \multicolumn{3}{c}{\textit{Dog}} 
        & \multicolumn{3}{c}{\textit{Sofa}}
        & \multicolumn{3}{c}{\textit{Bear}}
        & \multicolumn{3}{c}{\textit{Chair}} \\
        \cline{3-14}
    &   & PSNR$\uparrow$ & SSIM$\uparrow$ & LPIPS$\downarrow$ & PSNR$\uparrow$ & SSIM$\uparrow$ & LPIPS$\downarrow$ & PSNR$\uparrow$ & SSIM$\uparrow$ & LPIPS$\downarrow$ & PSNR$\uparrow$ & SSIM$\uparrow$ & LPIPS$\downarrow$\\
        \hline

\multirow{3}*{NeRF\cite{mildenhall2020nerf}} & LDR-OE  & 13.69   & 0.619   & 0.279   & 15.06  & 0.718  & 0.229  & 11.97 & 0.560 & 0.515    & 12.23   & 0.422  & 0.492  \\
                      & LDR-NE  &  ---    & ---   &  ---    & ---     &  ---      & ---      &  ---      & ---   &  ---      & ---   &   ---   & ---       \\
                      & HDR &  ---  & ---  &  ---      & ---     &  ---      & ---      &  ---      & ---   &  ---      & ---   &   ---   & ---      \\
        \hline
\multirow{3}*{NeRF-W\tnote{1} \cite{martin2021nerf}} & LDR-OE  & 31.01   & 0.967   & 0.022   & 30.76  & 0.955  & 0.029  & 32.24 & 0.978 & 0.021    & 28.01   & 0.840  & 0.161  \\
                      & LDR-NE  & 30.41   & 0.964   & 0.026   & 30.31  & 0.952  & 0.031  & 32.67 & 0.976 & 0.022   & 26.96   & 0.815  & 0.157  \\
                      & HDR &  ---    & ---   &  ---    & ---     &  ---      & ---      &  ---      & ---   &  ---      & ---   &   ---   & ---   \\
        \hline
\multirow{3}*{Ours} & LDR-OE  &\cellcolor{second}37.77 &\cellcolor{best}0.981 &\cellcolor{best}0.016 &\cellcolor{best}38.29 &\cellcolor{best}0.977 &\cellcolor{best}0.014 &\cellcolor{second}42.91 &\cellcolor{second}0.990 &\cellcolor{second}0.010 &\cellcolor{second}32.45 & \cellcolor{second}0.905 &\cellcolor{second}0.081  \\
                    & LDR-NE  &\cellcolor{second}36.52   &\cellcolor{second}0.976   &\cellcolor{second}0.018   &\cellcolor{second}38.35  &\cellcolor{second}0.976  &\cellcolor{best}0.014  &\cellcolor{second}41.19 &\cellcolor{second}0.987 &\cellcolor{second}0.012    &\cellcolor{second}30.78   &\cellcolor{second}0.886  &\cellcolor{second}0.083  \\
                    & HDR & \cellcolor{best}37.72   &\cellcolor{best}0.980   &\cellcolor{second}0.016   &\cellcolor{best}39.05  &\cellcolor{best}0.976  &\cellcolor{best}0.017  &\cellcolor{best}43.22 &\cellcolor{best}0.991 &\cellcolor{best}0.008  &\cellcolor{best}34.14   &\cellcolor{best}0.924  &\cellcolor{second}0.069  \\
        \hline
\multirow{3}*{NeRF-GT\tnote{2} \cite{mildenhall2020nerf}} & LDR-OE  &\cellcolor{best}38.43   &\cellcolor{best}0.981   &\cellcolor{second}0.017   &\cellcolor{second}37.91  &\cellcolor{second}0.975  &\cellcolor{second}0.046  &\cellcolor{best}43.84 &\cellcolor{best}0.991 &\cellcolor{best}0.009    &\cellcolor{best}33.79   &\cellcolor{best}0.926  &\cellcolor{best}0.070  \\
                      & LDR-NE  &\cellcolor{best}37.86   &\cellcolor{best}0.980   &\cellcolor{best}0.016   &\cellcolor{best}38.67  &\cellcolor{best}0.978  &\cellcolor{best}0.014  &\cellcolor{best}42.95 &\cellcolor{best}0.990 &\cellcolor{best}0.008 &\cellcolor{best}32.17  &\cellcolor{best}0.912  &\cellcolor{best}0.070  \\
                      & HDR & \cellcolor{second}35.66   &\cellcolor{second}0.967   &\cellcolor{best}0.007   &\cellcolor{second}36.38  &\cellcolor{second}0.955  &\cellcolor{second}0.044  &\cellcolor{second}38.43 &\cellcolor{second}0.971 &\cellcolor{second}0.014 &\cellcolor{second}33.72  &\cellcolor{second}0.922  &\cellcolor{best}0.010  \\
        \hline
    \end{tabular}
    \begin{tablenotes}
        \footnotesize
        \item[1] The exposures of input views for NeRF-W are randomly selected from all five exposures to learn five appearance vectors for testing.
        \item[2] A  version of  NeRF (as the upper bound of our method) that  is  trained  from  LDR  images  with  consistent exposures or HDR images.
     \end{tablenotes}
    \end{threeparttable}
    \label{tb:cmp_syn2}
    
\end{table*}

\begin{table*}[!th]
    \small
    \centering
    \caption{Quantitative comparisons with baseline methods on real scenes. LDR-OE denotes the average LDR results with exposure $t_1$, $t_3$, and $t_5$. LDR-NE denotes the average LDR results with exposure $t_2$, and $t_4$. We color code each column as \colorbox{best}{best} and \colorbox{second}{second best}.}
    
    \begin{threeparttable}
    \begin{tabular}{@{}p{49pt}|p{39pt}|p{20pt}<{\centering} p{20pt}<{\centering} p{20pt}<{\centering} p{20pt}<{\centering} p{20pt}<{\centering} p{20pt}<{\centering} p{20pt}<{\centering} p{20pt}<{\centering} p{20pt}<{\centering} p{20pt}<{\centering} p{20pt}<{\centering} p{20pt}<{\centering}}
        \hline 
        & & \multicolumn{3}{c}{\textit{Computer}} 
        & \multicolumn{3}{c}{\textit{Flower}}
        & \multicolumn{3}{c}{\textit{Luckycat}}
        & \multicolumn{3}{c}{\textit{Box}} \\
        \cline{3-14}
    &   & PSNR$\uparrow$ & SSIM$\uparrow$ & LPIPS$\downarrow$ & PSNR$\uparrow$ & SSIM$\uparrow$ & LPIPS$\downarrow$ & PSNR$\uparrow$ & SSIM$\uparrow$ & LPIPS$\downarrow$ & PSNR$\uparrow$ & SSIM$\uparrow$ & LPIPS$\downarrow$\\
        \hline

\multirow{2}*{NeRF\cite{mildenhall2020nerf}} 
& LDR-OE  & 14.68 & 0.697 & 0.281 & 14.60 & 0.504 & 0.524 & 13.67 & 0.706 & 0.262 & 17.06 & 0.770 & 0.233 \\
& LDR-NE  &  ---    & ---   &  ---    & ---     &  ---      & ---      &  ---      & ---   &  ---      & ---   &   ---   & ---       \\
\hline

\multirow{2}*{NeRF-W\tnote{1}\cite{martin2021nerf}} 
& LDR-OE  & 28.91 & 0.919 & 0.112 & 26.23 & 0.933 & 0.094 & 30.00 & 0.927 & 0.076 & 29.21 & 0.927 & 0.097 \\
& LDR-NE  & 27.54 & 0.892 & 0.136 & 26.84 & 0.939 & 0.078 & 30.78 & 0.940 & 0.058 & 29.59 & 0.923 & 0.104 \\
\hline

\multirow{2}*{Ours$\dag$} 
& LDR-OE  
& 31.41 & 0.944 & 0.086 
& 27.84 & 0.943 & 0.078 
& 31.82 & 0.937 & 0.067 
& 30.59 & 0.952 & 0.070\\
& LDR-NE  
& 29.01 & 0.923 & 0.112 
& 26.82 & 0.939 & 0.072 
& 31.40 & 0.944 & 0.059 
& 30.45 & \cellcolor{second}0.945 & \cellcolor{second}0.079\\
\hline

\multirow{2}*{Ours} 
& LDR-OE 
& \cellcolor{second}32.42 & \cellcolor{second}0.950 & \cellcolor{second}0.077 
& \cellcolor{second}29.81 & \cellcolor{second}0.948 & \cellcolor{second}0.069 
& \cellcolor{second}32.85 & \cellcolor{second}0.938 & \cellcolor{second}0.062 
& \cellcolor{second}31.54 & \cellcolor{second}0.953 & \cellcolor{second}0.068 \\
& LDR-NE 
& \cellcolor{second}31.21 & \cellcolor{second}0.931 & \cellcolor{second}0.098 
& \cellcolor{second}30.05 & \cellcolor{second}0.949 & \cellcolor{second}0.058 
& \cellcolor{second}33.13 & \cellcolor{second}0.948 & \cellcolor{second}0.051 
& \cellcolor{second}31.40 & 0.944 & \cellcolor{second}0.079 \\
\hline

\multirow{2}*{NeRF-GT\tnote{2}\cite{mildenhall2020nerf}} 
& LDR-OE  
& \cellcolor{best}34.34 & \cellcolor{best}0.955 & \cellcolor{best}0.075 
& \cellcolor{best}32.84 & \cellcolor{best}0.957 & \cellcolor{best}0.057 
& \cellcolor{best}34.56 & \cellcolor{best}0.951 & \cellcolor{best}0.049 
& \cellcolor{best}36.55 & \cellcolor{best}0.968 & \cellcolor{best}0.050 \\
& LDR-NE  
& \cellcolor{best}32.73 & \cellcolor{best}0.940 & \cellcolor{best}0.090
& \cellcolor{best}33.38 & \cellcolor{best}0.957 & \cellcolor{best}0.048 
& \cellcolor{best}36.42 & \cellcolor{best}0.962 & \cellcolor{best}0.035 
& \cellcolor{best}35.97 & \cellcolor{best}0.965 & \cellcolor{best}0.044 \\

        \hline
    \end{tabular}
    \begin{tablenotes}
        \footnotesize
        \item[1] The exposures of input views for NeRF-W are randomly selected from all five exposures to learn five appearance vectors for testing.
        \item[2] A  version of  NeRF (as the upper bound of our method) that  is  trained  from  LDR  images  with  consistent exposures.
        \item[$\dag$] An ablation study of our method that models the tone-mapping operations of RGB channels with a single MLP.
     \end{tablenotes}
    \end{threeparttable}
    \label{tb:cmp_real}
    
\end{table*}

\end{document}